\pdfoutput=1

\documentclass[11pt]{article}

\usepackage[]{EMNLP2023}

\usepackage{times}
\usepackage{latexsym}

\usepackage[T1]{fontenc}

\usepackage[utf8]{inputenc}

\usepackage{microtype}

\usepackage{inconsolata}

\usepackage{forest}
\usetikzlibrary{shadows}
\usepackage{amsmath}
\usepackage{amssymb}
\usepackage{cleveref}

\usepackage{booktabs}
\usepackage{siunitx}
\usepackage{etoolbox}
\usepackage{nicematrix}
\usepackage{multirow}
\usepackage{makecell}
\usepackage{enumitem}
\usepackage{xcolor,pifont}
\usepackage{fontawesome5}

\definecolor{lightcoral}{rgb}{0.94, 0.5, 0.5}
\definecolor{lightgreen}{rgb}{0.56, 0.93, 0.56}
\definecolor{harvestgold}{rgb}{0.85, 0.57, 0.0}
\definecolor{brightlavender}{rgb}{0.75, 0.58, 0.89}
\definecolor{capri}{rgb}{0.0, 0.75, 1.0}
\definecolor{carminepink}{rgb}{0.92, 0.3, 0.26}
\definecolor{celadon}{rgb}{0.67, 0.88, 0.69}
\definecolor{darkpastelgreen}{rgb}{0.01, 0.75, 0.24}

\newcommand*\circled[1]{\tikz[baseline=(char.base)]{
            \node[shape=circle,draw,inner sep=0.5pt,thick] (char) {#1};}}

\crefformat{section}{\S#2#1#3} 
\crefformat{subsection}{\S#2#1#3}
\crefformat{subsubsection}{\S#2#1#3}

\newrobustcmd{\B}{\bfseries}
\newcommand*\colourcheck[1]{%
  \expandafter\newcommand\csname #1check\endcsname{\textcolor{#1}{\ding{52}}}%
}
\newcommand*\colourcross[1]{%
  \expandafter\newcommand\csname #1cross\endcsname{\textcolor{#1}{\ding{55}}}%
}
\newcommand{\eg}{\emph{e.g.}}
\newcommand{\ie}{\emph{i.e.}}
\newcommand{\NA}{--}

\newcommand{\frozon}{\textcolor{capri}{\faIcon{snowflake}}}
\newcommand{\tuned}{\textcolor{carminepink}{\faIcon{fire}}}

\colourcheck{blue}
\colourcheck{green}
\colourcheck{darkpastelgreen}
\colourcross{blue}
\colourcross{applegreen}
\colourcross{red}

\title{How Do Large Language Models Capture the Ever-changing World Knowledge? A Review of Recent Advances}

\author{Zihan Zhang\textsuperscript{1}$^*$,  Meng Fang\textsuperscript{2}$^*$, Ling Chen\textsuperscript{1}, Mohammad-Reza Namazi-Rad\textsuperscript{3},
Jun Wang\textsuperscript{4} \\
        \textsuperscript{1}University of Technology Sydney \
        \textsuperscript{2}University of Liverpool \\
        \textsuperscript{3}University of Wollongong \
        \textsuperscript{4}University College London \\
        \texttt{Zihan.Zhang-5@student.uts.edu.au}, \texttt{Meng.Fang@liverpool.ac.uk} \\
        \texttt{Ling.Chen@uts.edu.au},
        \texttt{mrad@uow.edu.au},
        \texttt{junwang@cs.ucl.ac.uk} \\
        }

\begin{document}
\maketitle
\begin{abstract}

\def\thefootnote{*}\footnotetext{Equal contribution}\def\thefootnote{\arabic{footnote}}

Although large language models (LLMs) are impressive in solving various tasks, they can quickly be outdated after deployment. Maintaining their up-to-date status is a pressing concern in the current era. 
This paper provides a comprehensive review of recent advances in aligning LLMs with the ever-changing world knowledge without re-training from scratch.
We categorize research works systemically and provide in-depth comparisons and discussion. We also discuss existing challenges and highlight future directions to facilitate research in this field \footnote{We release the paper list at 
\url{https://github.com/hyintell/awesome-refreshing-llms} 
and will periodically update it.}.

\end{abstract}

\section{Introduction}
\label{introduction}

Large language models (LLMs) \citep{NEURIPS2020_1457c0d6, NEURIPS2022_b1efde53, chowdhery2022palm, zhang2022opt, openai2023gpt4, touvron2023llama, anil2023palm} trained on massive corpora from various sources (\eg, Wikipedia, Books, Github) implicitly store enormous amounts of world knowledge in their parameters \citep{petroni-etal-2019-language, roberts-etal-2020-much, jiang-etal-2020-know}, enabling them to act as versatile foundation models for performing various natural language processing (NLP) tasks  directly through in-context learning \citep{10.1145/3560815, openai2023gpt4, bubeck2023sparks, kamalloo2023evaluating} 
or for further fine-tuning for domain-specific uses \citep{singhal2022large, Med_PaLM_2, liu2023goat}.

\begin{figure}[t]
  \centering
  \includegraphics[width=\columnwidth]{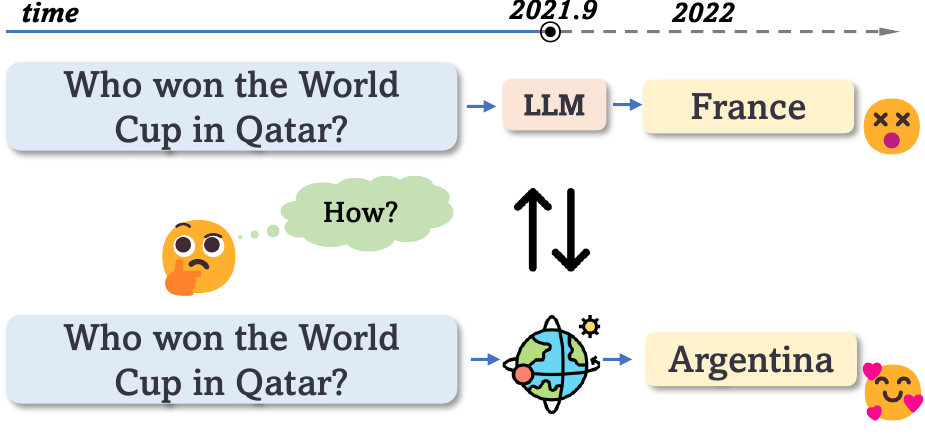}
  \caption{A trained LLM is static and can be outdated (\eg, ChatGPT; \citealt{OpenAI_chatgpt2022}). How can LLMs be aligned to the ever-changing world knowledge efficiently and effectively?}
  \label{fig_llm_align_world_example}
\end{figure}

Despite their impressive performance, LLMs are static after deployment, and there is no mechanism to update themselves or adapt to a changing environment \citep{kasai2022realtime, bubeck2023sparks}.
Our world, however, is dynamic and constantly evolving. As shown in Fig.\ref{fig_llm_align_world_example}, the static nature of trained LLMs makes the memorized knowledge quickly obsolete, which often causes hallucinations, rendering them unreliable for knowledge-intensive tasks \citep{lazaridou2022internetaugmented, luu-etal-2022-time, 10.1145/3571730, si2023prompting}.
In the era of LLMs, ensuring their alignment with the ever-changing world knowledge and maintaining their up-to-date status after deployment is a pressing concern because many users and downstream applications rely on them.
Unfortunately, simply re-training LLMs with the latest information is infeasible due to prohibitive costs \citep{patterson2021carbon}.

Intuitively, to update an LLM, one can either replace the obsolete knowledge stored \textit{implicitly} in the model with new ones by modifying its parameters, or override the outdated model outputs using new information \textit{explicitly} retrieved from the world.
Tremendous work has been proposed in the literature to implicitly or explicitly refresh deployed LLMs; however, these approaches, scattered among various tasks, have not been systematically reviewed and analyzed.

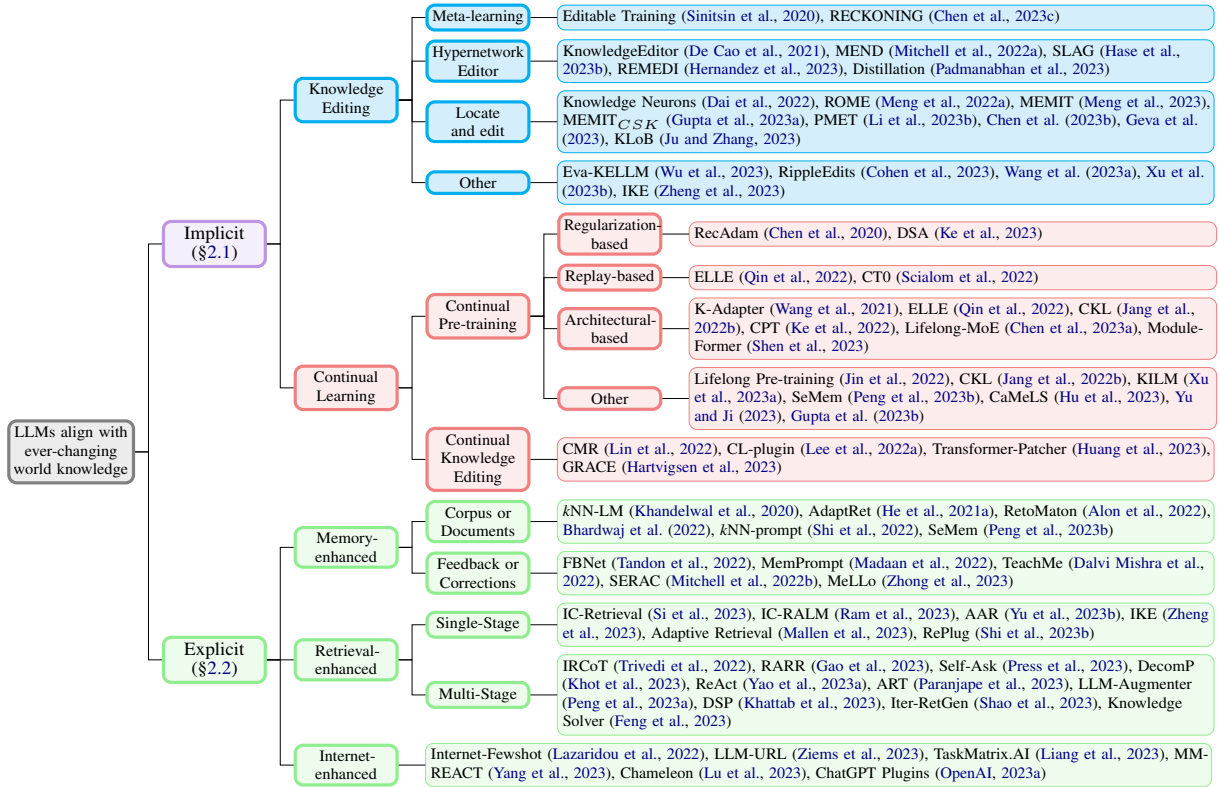
\begin{figure*}[ht]
\centering
\tikzset{
        my node/.style={
            draw,
            align=center,
            thin,
            text width=1.2cm, 
            rounded corners=3,
        },
        my leaf/.style={
            draw,
            align=left,
            thin,
            text width=8.5cm, 
            rounded corners=3,
        }
}
\forestset{
  every leaf node/.style={
    if n children=0{#1}{}
  },
  every tree node/.style={
    if n children=0{minimum width=1em}{#1}
  },
}
\begin{forest}
    for tree={%
        every leaf node={my leaf, font=\tiny},
        every tree node={my node, font=\tiny, l sep-=4.5pt, l-=1.pt},
        anchor=west,
        inner sep=2pt,
        l sep=10pt, 
        s sep=3pt, 
        fit=tight,
        grow'=east,
        edge={ultra thin},
        parent anchor=east,
        child anchor=west,
        if n children=0{tier=last}{},
        edge path={
            \noexpand\path [draw, \forestoption{edge}] (!u.parent anchor) -- +(5pt,0) |- (.child anchor)\forestoption{edge label};
        },
        if={isodd(n_children())}{
            for children={
                if={equal(n,(n_children("!u")+1)/2)}{calign with current}{}
            }
        }{}
    }
    [\tiny LLMs align with ever-changing world knowledge, draw=gray, color=gray!100, fill=gray!15, very thick, text=black, text width=1.5cm 
        [\scriptsize Implicit (\cref{implicitly_align}), color=brightlavender!100, fill=brightlavender!15, very thick, text=black
            [\tiny Knowledge Editing, color=cyan!100, fill=cyan!15, very thick, text=black
                [Meta-learning, color=cyan!100, fill=cyan!15, very thick, text=black
                    [ 
                        {
                         Editable Training \citep{Sinitsin2020Editable}, 
                         RECKONING \citep{chen2023reckoning}
                        },
                        color=cyan!100, fill=cyan!15, ultra thin, text=black, tier=E
                    ]
                ]
                [Hypernetwork Editor, color=cyan!100, fill=cyan!15, very thick, text=black
                    [ 
                        {
                        KnowledgeEditor \citep{de-cao-etal-2021-editing}, 
                        MEND \citep{mitchell2022fast}, 
                        SLAG \citep{hase-etal-2023-methods},
                        REMEDI \citep{hernandez2023inspecting},
                        Distillation \citep{padmanabhan2023propagating}
                        },
                        color=cyan!100, fill=cyan!15, ultra thin, text=black, tier=E
                    ]
                ]
                [Locate and edit, color=cyan!100, fill=cyan!15, very thick, text=black
                    [ 
                        {
                        Knowledge Neurons \citep{dai-etal-2022-knowledge}, 
                        ROME \citep{NEURIPS2022_6f1d43d5}, 
                        MEMIT \citep{meng2023massediting},
                        MEMIT$_{CSK}$ \citep{gupta2023editing},
                        PMET \citep{li2023pmet},
                        \citet{chen2023journey},
                        \citet{geva2023dissecting},
                        KLoB \citep{ju2023klob}
                        },
                        color=cyan!100, fill=cyan!15, ultra thin, text=black, tier=E
                    ]
                ]
                [Other, color=cyan!100, fill=cyan!15, very thick, text=black
                    [
                        {
                        Eva-KELLM \citep{wu2023evakellm}, 
                        RippleEdits \citep{cohen2023evaluating},
                        \citet{wang2023crosslingual},
                        \citet{xu2023language},
                        IKE \citep{zheng2023edit}
                        },
                        color=cyan!100, fill=cyan!15, ultra thin, text=black, tier=E
                    ]
                ]
            ]
            [\tiny Continual Learning, color=lightcoral!100, fill=lightcoral!15, very thick, text=black
                [\tiny Continual Pre-training, color=lightcoral!100, fill=lightcoral!15, very thick, text=black
                    [Regularization-based, color=lightcoral!100, fill=lightcoral!15, very thick, text=black
                        [
                            {
                            RecAdam \citep{chen-etal-2020-recall},
                            DSA \citep{ke2023continual}
                            },
                            color=lightcoral!100, fill=lightcoral!15, ultra thin, text=black, text width=6.8cm
                        ]
                    ]
                    [Replay-based, color=lightcoral!100, fill=lightcoral!15, very thick, text=black
                        [
                            {
                            ELLE \citep{qin-etal-2022-elle},
                            CT0 \citep{scialom-etal-2022-fine} 
                            },
                            color=lightcoral!100, fill=lightcoral!15, ultra thin, text=black, text width=6.8cm
                        ]
                    ]
                    [Architectural-based, color=lightcoral!100, fill=lightcoral!15, very thick, text=black
                        [
                            {
                            K-Adapter \citep{wang-etal-2021-k}, 
                            ELLE \citep{qin-etal-2022-elle},
                            CKL \citep{jang2022towards}, 
                            CPT \citep{ke-etal-2022-continual},
                            Lifelong-MoE \citep{pmlr-v202-chen23aq},
                            ModuleFormer \citep{shen2023moduleformer}
                            },
                            color=lightcoral!100, fill=lightcoral!15, ultra thin, text=black, text width=6.8cm
                        ]
                    ]
                    [Other, color=lightcoral!100, fill=lightcoral!15, very thick, text=black
                        [
                            {
                            Lifelong Pre-training \citep{jin-etal-2022-lifelong-pretraining}, 
                            CKL \citep{jang2022towards},
                            KILM \citep{xu2023kilm},
                            SeMem \citep{peng2023semiparametric},
                            CaMeLS \citep{hu2023metalearning}, 
                            \citet{yu2023self},
                            \citet{gupta2023continual}
                            },
                            color=lightcoral!100, fill=lightcoral!15, ultra thin, text=black, text width=6.8cm
                        ]
                    ]
                ]
                [Continual Knowledge Editing, color=lightcoral!100, fill=lightcoral!15, very thick, text=black
                    [
                        {
                        CMR \citep{lin-etal-2022-continual},
                        CL-plugin \citep{lee-etal-2022-plug},
                        Transformer-Patcher \citep{huang2023transformerpatcher},
                        GRACE \citep{hartvigsen2023aging}
                        },
                        color=lightcoral!100, fill=lightcoral!15, ultra thin, text=black, tier=E
                    ]
                ]
            ]
        ]
        [\scriptsize Explicit (\cref{explicitly_align}), color=lightgreen!100, fill=lightgreen!15, very thick, text=black 
            [Memory-enhanced, color=lightgreen!100, fill=lightgreen!15, very thick, text=black
                [Corpus or Documents, color=lightgreen!100, fill=lightgreen!15, very thick, text=black
                    [
                        {
                        \textit{k}NN-LM \citep{Khandelwal2020Generalization}, 
                        AdaptRet \citep{he-etal-2021-efficient},
                        RetoMaton \citep{pmlr-v162-alon22a},
                        \citet{bhardwaj2022adaptation},
                        \textit{k}NN-prompt \citep{shi-etal-2022-nearest},
                        SeMem \citep{peng2023semiparametric}
                        },
                        color=lightgreen!100, fill=lightgreen!15, ultra thin, text=black, tier=E
                    ]
                ]
                [Feedback or Corrections, color=lightgreen!100, fill=lightgreen!15, very thick, text=black
                    [
                        {
                        FBNet \citep{tandon-etal-2022-learning}, 
                        MemPrompt \citep{madaan-etal-2022-memory},
                        TeachMe \citep{dalvi-mishra-etal-2022-towards},
                        SERAC \citep{pmlr-v162-mitchell22a}, 
                        MeLLo \citep{zhong2023mquake}
                        },
                        color=lightgreen!100, fill=lightgreen!15, ultra thin, text=black, tier=E
                    ]
                ]
            ]
            [Retrieval-enhanced, color=lightgreen!100, fill=lightgreen!15, very thick, text=black
                [Single-Stage, color=lightgreen!100, fill=lightgreen!15, very thick, text=black
                    [
                        {
                        IC-Retrieval \citep{si2023prompting},
                        IC-RALM \citep{ram2023incontext},
                        AAR \citep{yu2023augmentationadapted},
                        IKE \citep{zheng2023edit},
                        Adaptive Retrieval \citep{mallen2023trust},
                        RePlug \citep{shi2023replug}
                        },
                        color=lightgreen!100, fill=lightgreen!15, ultra thin, text=black, tier=E
                    ]
                ]
                [Multi-Stage, color=lightgreen!100, fill=lightgreen!15, very thick, text=black
                    [
                        {
                        IRCoT \citep{trivedi2022interleaving},
                        RARR \citep{gao2023rarr},
                        Self-Ask \citep{press2023measuring},
                        DecomP \citep{khot2023decomposed},
                        ReAct \citep{yao2023react},
                        ART \citep{paranjape2023art},
                        LLM-Augmenter \citep{peng2023check},
                        DSP \citep{khattab2023demonstratesearchpredict},
                        Iter-RetGen \citep{shao2023enhancing},
                        Knowledge Solver \citep{feng2023knowledge}
                        },
                        color=lightgreen!100, fill=lightgreen!15, ultra thin, text=black, tier=E
                    ]
                ]
            ]
            [Internet-enhanced, color=lightgreen!100, fill=lightgreen!15, very thick, text=black
                [
                    {
                    Internet-Fewshot \citep{lazaridou2022internetaugmented},
                    LLM-URL \citep{ziems2023large},
                    TaskMatrix.AI \citep{liang2023taskmatrixai},
                    MM-REACT \citep{yang2023mmreact},
                    Chameleon \citep{lu2023chameleon},
                    ChatGPT Plugins \citep{OpenAI_plugin2023}
                    },
                    color=lightgreen!100, fill=lightgreen!15, ultra thin, text=black, tier=D, text width=10.2cm
                ]
            ]
        ]
    ]
\end{forest}
\caption{Taxonomy of methods to align LLMs with the ever-changing world knowledge (due to space limitation, please refer to Appendix \ref{append_complete_taxonomy} for a complete review).  
\textbf{Implicit} means the approaches seek to directly alter the knowledge stored in LLMs (\eg, parameters) (\cref{implicitly_align}), while \textbf{Explicit} means more often incorporating external resources to override internal knowledge (\eg, search engine) (\cref{explicitly_align}). 
}
\label{fig_taxonomy_of_methods}
\end{figure*}

In this review, we survey the recent compelling advances in aligning deployed LLMs with the ever-changing world knowledge.
We categorize research works systemically and highlight representative approaches in each category (\cref{section_taxonomy_of_methods}) and provide in-depth comparison with discussion for insights (\cref{section_comparison_and_discussion}).
Lastly, we discuss potential future directions to facilitate research in this field (\cref{section_future_directions}). 

\paragraph{Comparison with Related Work} 
To the best of our knowledge, surveys on this topic are scarce. 
Closest to our work, \citet{alkhamissi2022review} review pre-trained language models (LMs) as knowledge bases (KBs) and review a set of aspects that a LM should have to fully act as a KB; 
\citet{cao2023life} further divide the life cycle of knowledge in LLMs into five periods and survey how knowledge circulates; \citet{yao2023editing} conduct an empirical analysis of existing knowledge editing methods.
Despite partially overlapping with our discussion of knowledge editing in \cref{model_editing}, 
they only touch a subset of the scope that our survey studies and ignore other potentials in aligning LLMs with the world knowledge.
\citet{mialon2023augmented, wang2023interactive, qin2023tool}
study augmented, interactive, and tool learning of LLMs respectively, which share different goals from ours. 
Previous knowledge-enhanced LMs surveys \citep{zhu2021retrieving, wei2021knowledge, 10.1145/3512467, yin2022survey, zhen2022survey} focus on injecting knowledge into LMs, typically requiring modifying the model's architecture or re-training. 
Instead, we focus on the potential of how deployed LLMs capture the \textit{ever-changing} world knowledge effectively and efficiently without re-training.
\citet{wang2023comprehensive} provide a comprehensive review of forgetting in deep learning that is not limited to continual learning.
\citet{pan2023unifying} review potential approaches that unify knowledge graphs (KGs) and LLMs.
While structural knowledge, such as KGs, can broadly be categorised as explicit knowledge and augmented to LLMs for new knowledge, KG is static after creation and can still be outdated \citep{Ji_2022}.
New information or discoveries not yet incorporated into KGs may lead to outdated knowledge. 
However, how to efficiently update KGs is out of the scope of this survey.





\section{Taxonomy of Methods}
\label{section_taxonomy_of_methods}

Based on whether the method tends to directly alter the knowledge stored implicitly in LLMs, or leverage external resources to override the outdated knowledge, we roughly categorize them as \textit{implicit} (\cref{implicitly_align}) or \textit{explicit} (\cref{explicitly_align}) approaches. 
Fig.\ref{fig_taxonomy_of_methods}  provides a summary of representative works from each category
(See Fig.\ref{fig_full_taxonomy_of_methods} in Appendix for a complete review).
Detailed descriptions of the methods can be found in Appendix \ref{appendix_additional_approaches}.

\subsection{Implicitly Align LLMs with World Knowledge}
\label{implicitly_align}

Previous studies have shown that LLMs can implicitly memorize knowledge in their large number of parameters after being pre-trained on massive corpora \citep{petroni-etal-2019-language, roberts-etal-2020-much, jiang-etal-2020-know, singhal2022large}. 
To keep LLMs up-to-date and align with the current world knowledge, the straightforward way is to alter the model's behaviour from \textit{themselves} to generate desired outputs. 
Naively, one can regularly \textit{re-train} the model from scratch or \textit{fine-tune} the model with the latest corpora to align with current world knowledge.
However, re-training is expensive and environmentally unfriendly \citep{patterson2021carbon}, especially in the era of LLMs with billions of parameters. 
Fine-tuning without constraints may have a "butterfly effect" and affect other knowledge or skills present in the model \citep{doi:10.1073/pnas.1611835114, li2022large, alkhamissi2022review}.
To cope with this issue,
this line of work aims to design better strategies to modify the internal states of LLMs in a more controllable and efficient way, 
which can be categorized into
\textit{knowledge editing} (\cref{model_editing}) and \textit{continual learning} (\cref{continual_learning}).

\begin{figure*}[t]
  \centering
  \includegraphics[width=\textwidth]{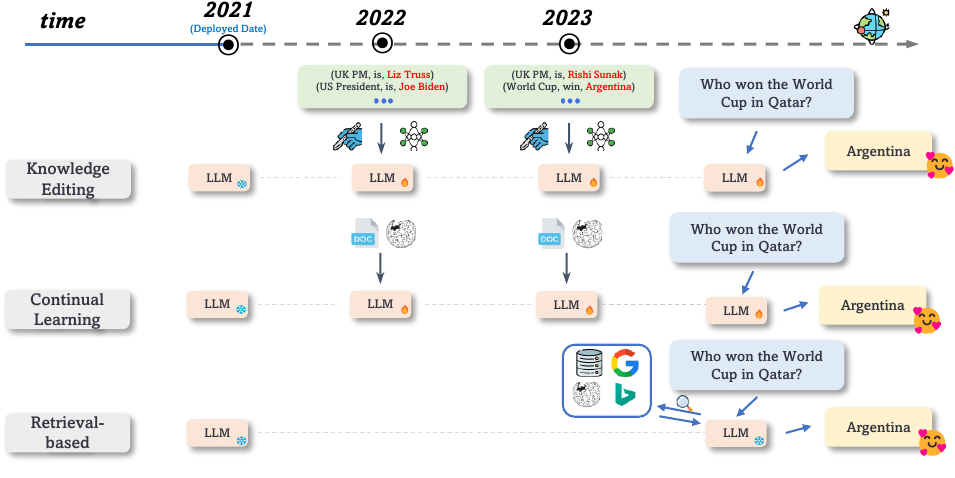}
  \caption{
  A high-level comparison of different approaches.
  }
  \label{fig_compare_methods}
\end{figure*}

\subsubsection{Knowledge Editing}
\label{model_editing}

Since tuning LLMs to learn new knowledge can be prohibitively expensive \citep{patterson2021carbon}, researchers seek efficient methods to directly update more specific, localized, or fine-grained knowledge that is preserved in LLMs \citep{mitchell2022fast}. 
Knowledge editing (KE) is an arising and promising research area that aims to alter the parameters of some specific knowledge stored in pre-trained models so that the model can make new predictions on those revised instances while keeping other irrelevant knowledge unchanged \citep{Sinitsin2020Editable, de-cao-etal-2021-editing, mitchell2022fast, NEURIPS2022_6f1d43d5, hase-etal-2023-methods, meng2023massediting}. 
In this section, we categorize existing methods into 
\textit{meta-learning}, \textit{hypernetwork}, and \textit{locate-and-edit} -based methods.

\paragraph{Meta-learning.}
This line of work generally focuses on the \textit{intrinsic} editability of the model itself, aiming to modify the model parameters so that they can be easily updated during inference \citep{de-cao-etal-2021-editing, mitchell2022fast}.
\citet{Sinitsin2020Editable} propose a model-agnostic meta-learning-based \citep{10.5555/3305381.3305498} method that trains neural networks in a way that the trained parameters can be easily edited afterwards. 
\citet{chen2023reckoning}
introduce a two-loop framework. In the inner training loop, they employ a few gradient updates to enable a pre-trained GPT-2 model \citep{radford2019language} to efficiently memorize external knowledge. Subsequently, in the outer loop, the model parameters are dynamically adjusted through optimal meta-parameter learning to incorporate additional knowledge that aids reasoning tasks. 

\paragraph{Hypernetwork Editor.}
In contrast to pre-modifying the pre-trained language model, an alternative approach in the field involves training \textit{extrinsic} editors that update knowledge during test time, thereby avoiding any modifications to the base model.
\citet{de-cao-etal-2021-editing} reframe editing the knowledge of a model as a \textit{learning-to-update} problem. Specifically, given a single data instance that needs to be updated, their trained hypernetwork \citep{ha2017hypernetworks} predicts a shift $\Delta \theta$ such that $\theta^{\prime}=\theta+\Delta \theta$, where $\theta$ is the original pre-trained LM weights and $\theta^{\prime}$ is the updated weights.
To keep editing effective while being easy to scale to larger LMs with billions of parameters, \citet{mitchell2022fast} decompose weight updates into low-rank components \citep{hu2022lora}, thus making it possible to scale to LLMs.
Orthogonal to \citet{mitchell2022fast}, \citet{hase-etal-2023-methods}  introduce a new training objective considering sequential, local, and generalizing model updates.
Although scaled beyond a single edit, their edit success rate significantly degrades when performing larger edits simultaneously.
Unlike the above methods that operate on the model's weight, \citet{hernandez2023inspecting} perform edits on the representation level.
\citet{padmanabhan2023propagating} employ knowledge distillation to transfer knowledge generated from a teacher model to a student model.

\paragraph{Locate and Edit.}
Generally, this line of work adopts the \textit{locate and edit} pattern: they first identify the location of specific knowledge stored in the model via different assumptions, then directly modify the weights or representations to update knowledge.
Inspired by the findings that feed-forward networks (FFN) in Transformer \citep{NIPS2017_3f5ee243} are key-value memories \citep{geva-etal-2021-transformer}, \citet{dai-etal-2022-knowledge} introduce the \textit{knowledge neurons} concept and propose a gradient-based knowledge attribution method to identify these knowledge neurons in FFNs. Further, without fine-tuning, they directly modify the corresponding value slots (\eg, embeddings) in the located knowledge neurons and successfully update or delete knowledge, demonstrating a preliminary potential to edit knowledge in LMs.

Different from \citet{geva-etal-2021-transformer}'s per-neuron view, \citet{NEURIPS2022_6f1d43d5} conduct casual tracing analysis on GPT-2 and hypothesize that the Transformer MLP can be viewed as a linear associative memory.
They verify their hypothesis by directly updating the middle-layer MLP weights with a rank-one update \citep{10.1007/978-3-030-58452-8_21}.
Following \citet{NEURIPS2022_6f1d43d5}'s work,
\citet{meng2023massediting} 
propose a scalable multi-layer method to update an LLM with thousands of facts simultaneously, significantly improves editing efficiency while maintaining generalization and specificity.
\citet{gupta2023editing} further adapt it to fix commonsense mistakes.
\citet{li2023pmet} find that Multi-Head Self-Attention (MHSA) weights do not require updating when introducing new knowledge. Based on this, they propose precisely updating FFN weights by simultaneously optimizing the Transformer component hidden states of MHSA and FFN to memorize target knowledge.
\citet{chen2023journey} propose an architecture-adapted multilingual integrated gradients method to localize knowledge neurons precisely across multiple architectures and languages.
\citet{geva2023dissecting} analyze the internal recall process of factual associations in auto-regressive LMs, opening new research directions for knowledge localization and model editing.

\paragraph{Other.} 

\citet{wu2023evakellm} propose an evaluation framework and dataset for measuring the effectiveness of knowledge editing of LLMs, as well as the ability to reason with the altered knowledge and cross-lingual knowledge transfer.
Similarly, \citet{cohen2023evaluating} evaluate the implications of an edit on related facts and show that existing methods fail to introduce consistent changes in the model's knowledge.
\citet{ju2023klob} propose an evaluation benchmark for locate-and-edit-based methods, aiming to reassess the validity of the locality hypothesis of factual knowledge.
\citet{wang2023crosslingual} and \citet{xu2023language} take multilingual into account and extend existing knowledge editing methods into cross-lingual scenarios.

\subsubsection{Continual Learning}
\label{continual_learning}


Continual learning (CL) aims to enable a model to learn from a continuous data stream across time while reducing catastrophic forgetting of previously acquired knowledge \citep{biesialska-etal-2020-continual}. With CL, a deployed LLM has the potential to adapt to the changing world without costly re-training from scratch \citep{bubeck2023sparks}.
In this section, we introduce approaches that employ CL for aligning LLMs with the current world knowledge, including
\textit{continual pre-training} and \textit{continual knowledge editing}.




\paragraph{Continual Pre-training.}
Unlike traditional continual learning, which sequentially fine-tunes a pre-trained LM on some specific downstream tasks (\eg, QA, text classification), \textit{continual pre-training} is used to further pre-train an LM to acquire new knowledge, where the data corpus is usually \textit{unsupervised} \citep{gururangan-etal-2020-dont, ke2023continua_surveyl}.
Since our target is the versatile foundation LLMs (\eg, GPT-4) that can be applied to many different use cases rather than a fine-tuned model designed for a specific task, we focus on the literature on continual pre-training. 

Early works \citep{gururangan-etal-2020-dont, rottger-pierrehumbert-2021-temporal-adaptation, lazaridou2021mind, dhingra-etal-2022-time} empirically analyze continuing LM pre-training on emerging domain or temporal data, showing the potential to update the base LM with new knowledge.
\citet{jang2022towards} explicitly categorize world knowledge as time-invariant, outdated, and new knowledge, which should be retained, acquired, and updated respectively by an LM when learning continually.
\citet{jin-etal-2022-lifelong-pretraining, jang-etal-2022-temporalwiki, jang2022towards} additionally implement traditional CL methods to alleviate \textit{catastrophic forgetting}, a phenomenon in which previously learned knowledge or abilities are degraded due to overwritten parameters \citep{doi:10.1073/pnas.1611835114}.
Among the literature, CL methods can be mainly categorized into 
\circled{1} \textbf{Regularization}, \circled{2} \textbf{Replay}, and
\circled{3} \textbf{Architectural}
-based methods.



\begin{table*}[!ht]
\scriptsize
\centering
\begin{NiceTabular}{llccccc}
\CodeBefore
  \rowcolor{gray!50}{1}
  \rowcolors{2}{gray!25}{white}
\Body
  \toprule 
  \textbf{Category} & \textbf{Representative Method} & \textbf{Base LM} & \textbf{LM Params} & \textbf{Augmentation} & \textbf{\makecell{No\\Training}} & \textbf{\makecell{Black\\-box}}  \\
  
   
  \midrule
  \Block{1-1}{\textbf{\makecell{Knowledge Editing}}} 
   & MEND \citep{mitchell2022fast} & T5 (11B) & \frozon & auxiliary model & \redcross & \redcross  \\
   & ROME \citep{NEURIPS2022_6f1d43d5} & GPT-J (6B) & \tuned & \NA & \darkpastelgreencheck & \redcross \\
   & CaliNET \citep{dong-etal-2022-calibrating} & T5 (0.7B) & \frozon & +params & \redcross & \redcross \\ 
   & MEMIT \citep{meng2023massediting} & \makecell{GPT-NeoX (20B)} & \tuned & \NA & \darkpastelgreencheck &  \redcross \\
    
  \midrule
  \Block{1-1}{\textbf{Continual Learning}} 
   & K-Adapter \citep{wang-etal-2021-k} & RoBERTa (0.3B) & \frozon & +params & \redcross & \redcross \\ 
   & CT0 \citep{scialom-etal-2022-fine} & T0 (3B) & \tuned & memory & \redcross & \redcross \\
   & DSA \citep{ke2023continual} & RoBERTa (0.1B) & \tuned & \NA & \redcross & \redcross  \\ 
   




   \midrule
   \Block{1-1}{\textbf{Memory-enhanced}} 
   
   & MemPrompt \citep{madaan-etal-2022-memory} & GPT-3 (175B) & \frozon & memory+retriever & \darkpastelgreencheck & \darkpastelgreencheck \\
   & SERAC \citep{pmlr-v162-mitchell22a} & T5 (0.7B) & \frozon & \makecell{memory\\+auxiliary model} & \redcross & \darkpastelgreencheck  \\
   & MeLLo \citep{zhong2023mquake} & \makecell{GPT-3.5 (175B)} &  \frozon & memory+retriever & \darkpastelgreencheck & \darkpastelgreencheck \\

   \midrule
   \Block{1-1}{\textbf{Retrieval-enhanced}} 

   & IRCoT \citep{trivedi2022interleaving} & GPT-3.5 (175B) & \frozon & \makecell{retriever} & \darkpastelgreencheck & \darkpastelgreencheck \\
   & RARR \citep{gao2023rarr} & PaLM (540B) & \frozon & \makecell{search engine\\+auxiliary model} & \darkpastelgreencheck & \darkpastelgreencheck \\
   & DecomP \citep{khot2023decomposed} & GPT-3 (175B) & \frozon & \makecell{retriever} & \darkpastelgreencheck & \darkpastelgreencheck \\
   & ReAct \citep{yao2023react} & PaLM (540B) & \frozon & \makecell{search engine} & \darkpastelgreencheck & \darkpastelgreencheck \\
   & FLARE \citep{jiang2023active} & GPT-3.5 (175B) & \frozon & \makecell{retriever/search engine} & \darkpastelgreencheck & \darkpastelgreencheck \\

   \midrule
   \Block{1-1}{\textbf{Internet-enhanced}} 
   & \citet{lazaridou2022internetaugmented} & Gopher (280B) & \frozon & \makecell{search engine} & \darkpastelgreencheck & \darkpastelgreencheck \\
    & CRITIC \citep{gou2023critic} & GPT-3.5 (175B) & \frozon & \makecell{various tools} & \darkpastelgreencheck & \darkpastelgreencheck \\
    & Chameleon \citep{lu2023chameleon} & GPT-4 (?B) & \frozon & \makecell{various tools} & \darkpastelgreencheck & \darkpastelgreencheck \\
  
  \bottomrule 
\end{NiceTabular}
    \caption{Comparison between representative methods (refer to Appendix \ref{append_complete_taxonomy} for a complete review). \tuned \ means the parameters of the original LM are modified, while \frozon \ means they are unchanged; 
    \textbf{Augmentation} means additional components used;
    \textbf{No Training} indicates the method does not require additional training; 
    \textbf{Black-box} refers to whether the method suits non-publicly available models (\eg, no model architecture, parameters, activations, or gradients are available).
    }
\label{table_comparison_methods}
\end{table*}

\circled{1} \textbf{Regularization}.
To mitigate forgetting, regularization-based methods apply regulations to penalize the changes of the critical parameters learned from previous data.
\citet{chen-etal-2020-recall} improve the traditional EWC \citep{doi:10.1073/pnas.1611835114} by recalling previously learned knowledge through the pre-trained parameters, and the method continually learns new information using a multi-task learning objective.
\citet{ke2023continual} compute the importance of each unit (\ie, attention head and neuron) to 
the general knowledge in the LM using a proxy based on model robustness to preserve learned knowledge. When continually learning new domains, the approach prevents catastrophic forgetting of the general and domain knowledge and encourages knowledge transfer via soft-masking and contrastive loss.

\circled{2} \textbf{Replay}.
These methods generally reduce forgetting by replaying previous training data when learning new data.
Assuming that the initial pre-training corpus is available, 
\citet{he-etal-2021-analyzing} use a gradual decay mix-ratio to adjust the quantity of the pre-training corpus mixed in the new data when learning sequentially.
ELLE \citep{qin-etal-2022-elle} and CT0 \citep{scialom-etal-2022-fine} also mix the old data while learning new data. However, ELLE starts the pre-training from a newly initialized and relatively small BERT \citep{devlin-etal-2019-bert} and GPT \citep{radford2018improving}, while CT0 continues learning from T0-3B \citep{sanh2022multitask}, a pre-trained and instruction-tuned model.

\circled{3} \textbf{Architectural}.
These methods typically alleviate forgetting by using different subsets of parameters for distinct tasks or domains.
\citet{wang-etal-2021-k, hu2022lora, ke-etal-2022-continual}
freeze the original parameters of the LM to preserve the learned knowledge and add lightweight tunable parameters for continually learning new knowledge. 
\citet{wang-etal-2021-k} add separate adapters \citep{houlsby2019parameter} for each new task, while \citet{ke-etal-2022-continual} let all domains share adapters and employ task masks to protect critical neurons from being updated.
DEMix-DAPT \citep{gururangan-etal-2022-demix} replaces every FFN layer in Transformer with a separate domain expert mixture layer, containing one expert per domain. 
When learning new knowledge, they only train the newly added expert in each DEMix layer while fixing all other experts.
Similarly, Lifelong-MoE \citep{pmlr-v202-chen23aq} progressively expands experts 
to increase model capacity for learning new knowledge, and mitigates forgetting by freezing previously trained experts and gatings with output-level regularization.
\citet{qin-etal-2022-elle} enlarge the model's width and depth to attain learning efficiency and employ memory replay to reduce forgetting.


\circled{4} \textbf{Other Methods}.
\citet{hu2023metalearning} meta-trains an importance-weighting model to reweight the per-token loss of the continual data stream, intending to quickly adapt the base LM to new knowledge.
\citet{peng2023semiparametric} apply \textit{k}NN-LM \citep{Khandelwal2020Generalization} to continual learning from streaming data and selectively store hard cases in a non-parametric memory, significantly improving the data-wise and model-wise scalability.
\citet{yu2023self} assess self-information-update in LLMs via CL and mitigate exposure bias by incorporating the selection of relevant facts into training losses.



\paragraph{Continual Knowledge Editing.}
\citet{lin-etal-2022-continual, lee-etal-2022-plug, huang2023transformerpatcher} and \citet{ hartvigsen2023aging} propose a more realistic setting that a deployed LM should be constantly corrected to fix its prediction errors, showing the potential to align the model with the latest world knowledge.
\citet{lin-etal-2022-continual} benchmark the continual model refinement problem by implementing traditional CL methods.
\citet{lee-etal-2022-plug} and \citet{ hartvigsen2023aging} freeze the LM's original parameters and continually introduce trainable neurons to the FFN layer to rectify problematic model behaviors. 
In contrast, \citet{hartvigsen2023aging} learn to cache a chosen layer's activations in a key-value-based codebook and retrieve activations when previous similar edits have been performed. Without influencing unrelated inputs, it can efficiently edit the model thousands of times in a row while generalizing edits to previously unseen inputs.

\subsection{Explicitly Align LLMs with World Knowledge}
\label{explicitly_align}

Although altering the knowledge implicitly stored in LLMs has shown to be effective \citep{jang2022towards, meng2023massediting}, it remains unclear whether it will affect the models' general abilities due to the complexity of neural networks.
In contrast, explicitly augmenting LLMs with the latest information retrieved from various sources can effectively adapt the models to new world knowledge without  affecting the original LLMs \citep{mialon2023augmented}.
However, previous retrieval-augmented methods \citep{karpukhin-etal-2020-dense, guu2020retrieval, lewis2020retrieval, izacard2022atlas, borgeaud2022improving, jiang-etal-2022-retrieval, kaur-etal-2022-lm} 
usually jointly train a retriever and an LM in an end-to-end fashion, 
making it challenging to apply to a deployed LLM (\eg, GPT-3).
Recently, researchers have focused on equipping a fixed LLM with external memory (\textit{memory-enhanced}; \cref{memory_enhanced}), an off-the-shelf retriever (\textit{retrieval-enhanced}; \cref{retrieval_enhanced}), or Internet (\textit{Internet-enhanced}; \cref{internet_enhanced}) to cope with this issue.

\subsubsection{Memory-enhanced Methods}
\label{memory_enhanced}

Pairing a static LLM with a growing non-parametric memory enables it to capture information beyond its memorized knowledge during inference \citep{wu2022memorizing}. The external memory can store a recent 
\textit{corpus} or \textit{feedback} 
that contains new information to guide the model generation.


\paragraph{Storing Corpus or Documents.}
\textit{k}NN-LM \citep{Khandelwal2020Generalization} stores every \texttt{<context, token>} as key-value pairs from a corpus in memory. During inference, it calculates the probability of the next token by interpolating a fixed LM with a distribution retrieved from the \textit{k} nearest tokens in the memory. 
Following this vein, \citet{he-etal-2021-efficient, drozdov-etal-2022-cant, pmlr-v162-alon22a} improve the efficiency of \textit{k}NN-LM by skipping unnecessary retrieval. 
\citet{meng2022gnnlm} build an additional graph neural network to aggregate information from the retrieved context for better generation.
\citet{peng2023semiparametric} improve the scalability of \textit{k}NN-LM for continual learning, while \citet{shi-etal-2022-nearest} apply it for zero-shot inference on downstream tasks.


\paragraph{Storing Feedback or Corrections.}
Inspired by the fact that humans can learn from past mistakes, this line of work stores user feedback in memory to fix the model's problematic predictions and avoids similar errors in the future.
By querying the memory, the base LLM gains \textit{editability} to update its outdated knowledge.
\citet{kassner-etal-2021-beliefbank, tandon-etal-2022-learning} train an auxiliary corrector to apply feedback to repair the model output. 
\citet{dalvi-mishra-etal-2022-towards} allow users to interact with the system to check its facts and reasoning and correct it when it is wrong.
Similarly, \citet{madaan-etal-2022-memory} equip GPT-3 with a growing memory, where the key is a misunderstanding question, and the value is the corrective feedback. 
Instead of storing user feedback, \citet{pmlr-v162-mitchell22a, zhong2023mquake} explicitly preserve updated knowledge in memory. Given an input, \citet{pmlr-v162-mitchell22a} first apply a classifier to determine if a relevant edit exists in the memory and perform knowledge updating through a counterfactual model. 
Conversely, \citet{zhong2023mquake} decompose complex questions and ask the base model to generate a temporary answer. They revise the model output when the generated answer contradicts the retrieved facts from memory.

\begin{figure}[t]
  \centering
  \includegraphics[width=\columnwidth]{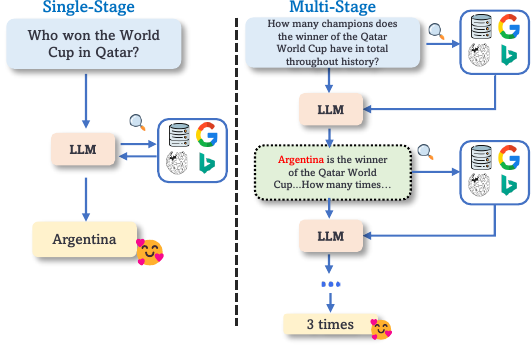}
  \caption{
  \textbf{Single-Stage} (left) typically retrieves once, while \textbf{Multi-Stage} (right) involves multiple retrievals or revisions to solve complex questions (\cref{retrieval_enhanced}).
  }
  \label{fig_compare_single_multi_stage}
\end{figure}

\subsubsection{Retrieval-enhanced Methods}
\label{retrieval_enhanced}

Leveraging an off-the-shelf retriever and the in-context learning ability of LLMs \citep{NEURIPS2020_1457c0d6}, this line of work designs better retrieval strategies to incorporate world knowledge into a fixed LLM through prompting, which can be divided into 
\textit{single-stage} and \textit{multi-stage} (Fig.\ref{fig_compare_single_multi_stage}).


\paragraph{Single-Stage.}
To ground the model with external knowledge during generation, \citet{ram2023incontext, si2023prompting} adopt zero-shot and few-shot retrieval respectively and directly prepend the retrieved documents to the input without changing the base LLM. 
\citet{zheng2023edit} retrieve similar edit demonstrations for each input and perform in-context knowledge editing. Compared with gradient-based knowledge editing (\cref{model_editing}), they have competitive editing performance with fewer side effects.
Arguing that the general-purpose retrievers could be sub-optimal, \citet{yu2023augmentationadapted} adopt a small source LM to provide LM-preferred signals to train an adaptive retriever.
\citet{mallen2023trust} employ a heuristic based on entity popularity and only retrieve relevant context when the input questions are less popular, which improves performance and reduces inference costs.
Unlike above, to address the limited model's context length, \citet{shi2023replug} prepend each retrieved document separately to an LLM and then ensemble output probabilities from different passes.



\paragraph{Multi-Stage.}
When solving complex questions, retrieving information only once based on the input is often inadequate. This branch of work aims to transform single-stage retrieval into multi-stage retrieval in order to solve complex tasks, usually by leveraging reasoning.
\citet{trivedi2022interleaving} interleave knowledge retrieval with chain-of-thoughts (CoT; \citealt{wei2022chain}) generation to solve complex multi-step reasoning questions. 
Similarly, \citet{press2023measuring, khot2023decomposed, yao2023react, jiang2023active, shao2023enhancing} decompose questions into sub-questions to provide a specific context for retrieval with model generation.
\citet{paranjape2023art, chen2023chatcot, inaba2023multitoolcot} further enable the usage of different tools to solve various tasks.
Unlike the simple \textit{retrieve-then-read} paradigm, \citet{khattab2023demonstratesearchpredict} pass intermediate messages between an LLM and a retriever;  
\citet{gao2023rarr, he2022rethinking, zhao2023verifyandedit, yu2023improving} retrieve after generation and perform post-edit revisions for more faithful outputs.
\citet{peng2023check} iteratively revise ChatGPT to improve model responses using feedback and external knowledge.
\citet{feng2023knowledge} teach LLMs themselves to search for knowledge from external knowledge graphs (KGs) via prompting and simplify searching into a multi-hop decision sequence, allowing explainable decision-making of the processes.

\subsubsection{Internet-enhanced Methods}
\label{internet_enhanced}

Prior retrieval-augmented work relies on \textit{static} or \textit{offline} knowledge sources (\eg, Wikipedia dump), which may not be sufficiently up-to-date or complete for tasks that require the latest knowledge  \citep{kasai2022realtime, zhang2023interpretable, li2023web}.
A recent trend uses the whole web as the knowledge source and equips LLMs with the Internet to support real-time information seeking \citep{nakano2022webgpt, menick2022teaching, komeili-etal-2022-internet, shuster2022blenderbot, qin2023webcpm, liu2023evaluating}.
\citet{lazaridou2022internetaugmented} augment few-shot QA prompting with the context retrieved from Google search.
\citet{press2023measuring, jiang2023active} interleave reasoning with web search. 
Recently, tools such as LangChain \citep{langchain2022} and ChatGPT Plugins \citep{OpenAI_plugin2023} connect a deployed LLM to the Internet without training, making them more powerful for solving knowledge-intensive tasks.
Beyond search engines, \citet{yao2023react, liang2023taskmatrixai, paranjape2023art, yang2023mmreact, gou2023critic, lu2023chameleon} treat LLMs as central planners and compose various plug-and-play tools for solving complex questions.

\section{Comparison and Discussion}
\label{section_comparison_and_discussion}

We present the comparison of different methods in Table \ref{table_comparison_methods} and in Fig.\ref{fig_compare_methods},
and the characteristics of different methods in Table \ref{table_comparison_characteristics} in Appendix.

\paragraph{Discussion of Implicit Methods (\cref{implicitly_align}).} 
Compared to naive re-training or fine-tuning, KE and CL can effectively update obsolete knowledge in LLMs while minimizing interference on irrelevant ones.
We identify their major differences: 
\circled{1} \textbf{Scale}.
Existing KE methods focus on updating small-scale and localized knowledge, typically on synthetic fact pairs \citep{mitchell2022fast, NEURIPS2022_6f1d43d5}.
While one can perform thousands of edits simultaneously \citep{meng2023massediting}, updating enormous knowledge in LLMs may be cumbersome.
In contrast, CL enhances models' adaptability via tuning larger-scale parameters, thus updating more knowledge at scale \citep{jang2022towards}.
However, KE provides fine-grained controllability when specific knowledge needs to be altered, which is unachievable by CL;
\circled{2} \textbf{Forgetting}. 
Applying KE methods on LLMs frequently in response to the ever-changing world is sub-optimal due to catastrophic forgetting \citep{huang2023transformerpatcher, hartvigsen2023aging}; CL mitigates this issue when learning new knowledge;
\circled{3} \textbf{Cost}. 
CL is generally more computationally expensive than KE due to larger-scale weight updating.

\paragraph{Discussion of Explicit Methods (\cref{explicitly_align}).} 
Explicit methods use new knowledge retrieved from the world to override old knowledge in an LLM during generation.
Despite being effective, memory- and retrieval-enhanced methods must periodically maintain the external memory and the knowledge sources in response to the ever-changing world \citep{kasai2022realtime}. Conversely, Internet-enhanced methods enable real-time knowledge seeking, although potentially suffering from noisy and low-quality web content \citep{li2023web, luo2023sail}.
Compared to single-stage retrieval, multi-stage retrieval can solve more complex problems. Nevertheless, they may interrupt the generation with multiple  retrievals or revisions, leading to considerable inference overheads \citep{shao2023enhancing}.

\paragraph{Updating LLMs Implicitly or Explicitly?} 
We observe an increasing trend of explicitly aligning LLMs with world knowledge while keeping the model untouched (Table \ref{table_full_comparison_methods} in Appendix).
Compared to explicit approaches:
\circled{1} \textbf{Applicability}.
Implicit methods usually require modifying LLM's parameters or gradients, making it challenging to update closed-source models;
\circled{2} \textbf{Side Effects}.
Although constraints have been added to avoid editing irrelevant knowledge \citep{mitchell2022fast, meng2023massediting} or forgetting general knowledge \citep{jang2022towards}, modifying the LLM's parameters inevitably has side effects that may hurt the performance, which is hard to estimate due to the complexity of neural networks \citep{brown2023edit};
\circled{3} \textbf{Efficiency}.
Implicit methods typically require training, while most explicit methods leverage a fixed LLM and an off-the-shelf retriever, erasing the necessity of training.
However, explicit methods do not directly modify the intrinsic knowledge within LLMs, but instead rely on on-the-fly retrieval during inference, resulting in a notable increase in the computational cost of inference.


\section{Challenges and Future Directions}
\label{section_future_directions}

\paragraph{Robust and Efficient Knowledge Editing.}
KE offers fine-grained knowledge updating, which is desirable in some scenarios.
Despite promising, KE is still in its infancy stage. 
\circled{1} \textbf{Various knowledge}.
It is challenging to renew the internal knowledge stored in the parameters of LLMs, and 
existing efforts have only explored updating relational knowledge while overlooking other knowledge \citep{meng2023massediting};
\circled{2} \textbf{Edit dataset}.
Current KE methods assume edited knowledge pairs exist, which must be annotated beforehand. In reality, how do LLMs know what knowledge is outdated and thus needs to be updated \citep{zhang2023mitigating, yin2023large}?
\circled{3} \textbf{Memorization mechanism}.
\citet{hase2023does} argue that the localization of specific knowledge via casual tracing may not be reliable, calling for a better understanding of the internal memorization of LLMs \citep{NEURIPS2022_fa0509f4, carlini2023quantifying};
\circled{4} \textbf{Generalization}.
Recent studies \citep{onoe2023lms, zhong2023mquake} find that existing KE methods show little propagation of edited knowledge, meaning the LLM cannot make further reasoning based on the newly acquired knowledge;
\circled{5} \textbf{Effectiveness and efficiency}.
Although early efforts have been made \citep{hernandez2023inspecting, huang2023transformerpatcher, hartvigsen2023aging}, methods to effectively, efficiently, 
and continually renew the knowledge of LLMs at scale have yet to be thoroughly explored.

\paragraph{Efficient Continual Learning of LLMs.} 
A continual pre-trained LLM can update its internal knowledge and adapt to the changing world, but maintaining the general knowledge required for downstream tasks without forgetting is challenging \citep{ke2023continua_surveyl}.
Moreover, existing methods are limited to small-scale LMs, leaving CL of LLMs rarely studied.
While parameter-efficient tuning \citep{ding2022delta} may be beneficial, 
it remains under-explored to align an LLM with the dynamic world via CL.

\paragraph{Solving Knowledge Conflicts.} 
Replacing old knowledge with new ones can cause knowledge conflicts regardless of using implicit or explicit methods. 
For implicit methods, these side effects are only evaluated in specific settings, and there is no idea of how the general skills of LLMs are impacted \citep{brown2023edit}.
For retrieval-based methods, knowledge retrieved from the world can contradict the knowledge memorized inside LLMs, and LLMs sometimes favour their internal knowledge rather than the provided context during generation (an example in Fig.\ref{fig_chatgpt_knowledge_conflict}; \citealt{neeman2022disentqa, li2022large, chen-etal-2022-rich}). 
While initial attempts have been made \citep{mallen2023trust, zhou2023contextfaithful, xie2023adaptive}, they are still limited.

\paragraph{Robust and Efficient Retrieval.} 
Interacting with external resources can cause interruptions during generation, significantly increasing inference overheads, especially for multi-stage methods that involve multiple retrievals or revisions.
Potential remedies may be efficient memory management \citep{peng2023semiparametric, kang2023think, cheng2023decouple} or selective retrieval that only consults external resources when necessary \citep{mallen2023trust}.
On the other hand, the retrieved context can be irrelevant and noisy, which may distract LLMs \citep{shi2023large, luo2023sail}, or too long, which exceeds the input limits and renders high cost \citep{shi2023replug}.






\paragraph{Comprehensive Evaluation and Benchmarks.} 
Although approaches of different categories can align the trained LLMs with the changing world without re-training, their effectiveness is primarily evaluated on synthetic datasets in specific settings, which might not be comprehensive \citep{jang-etal-2022-temporalwiki, jang2022towards, hoelscherobermaier2023detecting}.
Moreover, although efforts have been made to evaluate KE \citep{wu2023evakellm, cohen2023evaluating, ju2023klob},
there is no quantitative comparison of methods of different categories (\ie, comparing KE vs. CL vs. retrieval-based methods), hindering their application in different scenarios.
Lastly, existing benchmarks are too \textit{static} to measure the dynamic world, which calls for real-time evaluation benchmarks \citep{liška2022streamingqa, kasai2022realtime}.







\section{Conclusion}

In this paper, we systematically review recent advances in aligning LLMs with the ever-changing world knowledge without re-training. 
We summarize existing approaches and categorize them based on whether they tend to directly alter the knowledge stored implicitly in LLMs, or leverage external resources to override the outdated knowledge.
We comprehensively compare methods of different categories and point out challenges and future directions to facilitate research in this area.

\section*{Limitations}

In this paper, we systematically review recent advances in aligning LLMs with the ever-changing world knowledge without re-training. 
We compare our work with the related surveys in \cref{introduction} and will periodically add related approaches.
Despite our best efforts, there exist some limitations in this paper:

\paragraph{Scope.} 
In this survey, we do not review knowledge-enhanced approaches that require re-training because we focus on the already trained (deployed) models and how to keep them up-to-date.
We refer interested readers to the relevant knowledge-enhanced LMs surveys \citep{zhu2021retrieving, wei2021knowledge, 10.1145/3512467, yin2022survey, zhen2022survey}.
Second, in terms of world knowledge, we focus on text-based knowledge and leave other kinds of knowledge, such as images, video, audio, etc., and structural knowledge, such as knowledge graphs (KGs) and databases, for future work.
Third, we mainly review the cutting-edge approaches within three years (mostly in 2022 and 2023) in \cref{section_taxonomy_of_methods}, mainly from the ACL, EMNLP, NAACL, TACL, NeurIPS, ICML, ICLR, arXiv. 
Despite our best efforts, by no means the surveyed methods are complete, and we may miss some important references. 
Lastly, we cannot afford all the technical details due to page limitations and may only provide brief introductions. We provide additional discussion of approaches in Appendix \ref{appendix_additional_approaches}.

\paragraph{Taxonomy.} 
It should be noted that some approaches are hybrid and can be categorized into different branches. 
We mainly categorize them based on their main components or mechanism.
For instance, all methods in \cref{explicitly_align} require retrieving from external resources. Memory-enhanced methods (\cref{memory_enhanced}) pay more attention to the design of external memory, while paying little attention to retrieval strategies.

\paragraph{Empirical Comparison.} 
We provide detailed comparisons and discussions in \cref{section_comparison_and_discussion} and potential future directions in \cref{section_future_directions}. All the conclusions are proposed based on empirical summarization of existing works. However, as the field evolves fast, these empirical conclusions might be inapplicable. 
We will update the latest opinions timely. 
In addition, we do not provide quantitative comparisons through experiments since there is no unified evaluation benchmarks of different categories. 
Quantitative evaluation (benchmarks) is a challenging and interesting future direction to fairly compare methods of different categories to align LLMs with updated world knowledge (\cref{section_future_directions}). 
We will leave quantitative comparisons and analysis as future work.


\section*{Acknowledgements}

This work is supported by TPG Telecom. 
We would like to thank anonymous reviewers for their valuable comments. 

\bibliography{anthology,custom}

\begin{thebibliography}{181}
\expandafter\ifx\csname natexlab\endcsname\relax\def\natexlab#1{#1}\fi

\bibitem[{AlKhamissi et~al.(2022)AlKhamissi, Li, Celikyilmaz, Diab, and Ghazvininejad}]{alkhamissi2022review}
Badr AlKhamissi, Millicent Li, Asli Celikyilmaz, Mona Diab, and Marjan Ghazvininejad. 2022.
\newblock \href {http://arxiv.org/abs/2204.06031} {A review on language models as knowledge bases}.

\bibitem[{Alon et~al.(2022)Alon, Xu, He, Sengupta, Roth, and Neubig}]{pmlr-v162-alon22a}
Uri Alon, Frank Xu, Junxian He, Sudipta Sengupta, Dan Roth, and Graham Neubig. 2022.
\newblock \href {https://proceedings.mlr.press/v162/alon22a.html} {Neuro-symbolic language modeling with automaton-augmented retrieval}.
\newblock In \emph{Proceedings of the 39th International Conference on Machine Learning}, volume 162 of \emph{Proceedings of Machine Learning Research}, pages 468--485. PMLR.

\bibitem[{Anil et~al.(2023)Anil, Dai, Firat, Johnson, Lepikhin, Passos, Shakeri, Taropa, Bailey, Chen, Chu, Clark, Shafey, Huang, Meier-Hellstern, Mishra, Moreira, Omernick, Robinson, Ruder, Tay, Xiao, Xu, Zhang, Abrego, Ahn, Austin, Barham, Botha, Bradbury, Brahma, Brooks, Catasta, Cheng, Cherry, Choquette-Choo, Chowdhery, Crepy, Dave, Dehghani, Dev, Devlin, Díaz, Du, Dyer, Feinberg, Feng, Fienber, Freitag, Garcia, Gehrmann, Gonzalez, Gur-Ari, Hand, Hashemi, Hou, Howland, Hu, Hui, Hurwitz, Isard, Ittycheriah, Jagielski, Jia, Kenealy, Krikun, Kudugunta, Lan, Lee, Lee, Li, Li, Li, Li, Li, Lim, Lin, Liu, Liu, Maggioni, Mahendru, Maynez, Misra, Moussalem, Nado, Nham, Ni, Nystrom, Parrish, Pellat, Polacek, Polozov, Pope, Qiao, Reif, Richter, Riley, Ros, Roy, Saeta, Samuel, Shelby, Slone, Smilkov, So, Sohn, Tokumine, Valter, Vasudevan, Vodrahalli, Wang, Wang, Wang, Wang, Wieting, Wu, Xu, Xu, Xue, Yin, Yu, Zhang, Zheng, Zheng, Zhou, Zhou, Petrov, and Wu}]{anil2023palm}
Rohan Anil, Andrew~M. Dai, Orhan Firat, Melvin Johnson, Dmitry Lepikhin, Alexandre Passos, Siamak Shakeri, Emanuel Taropa, Paige Bailey, Zhifeng Chen, Eric Chu, Jonathan~H. Clark, Laurent~El Shafey, Yanping Huang, Kathy Meier-Hellstern, Gaurav Mishra, Erica Moreira, Mark Omernick, Kevin Robinson, Sebastian Ruder, Yi~Tay, Kefan Xiao, Yuanzhong Xu, Yujing Zhang, Gustavo~Hernandez Abrego, Junwhan Ahn, Jacob Austin, Paul Barham, Jan Botha, James Bradbury, Siddhartha Brahma, Kevin Brooks, Michele Catasta, Yong Cheng, Colin Cherry, Christopher~A. Choquette-Choo, Aakanksha Chowdhery, Clément Crepy, Shachi Dave, Mostafa Dehghani, Sunipa Dev, Jacob Devlin, Mark Díaz, Nan Du, Ethan Dyer, Vlad Feinberg, Fangxiaoyu Feng, Vlad Fienber, Markus Freitag, Xavier Garcia, Sebastian Gehrmann, Lucas Gonzalez, Guy Gur-Ari, Steven Hand, Hadi Hashemi, Le~Hou, Joshua Howland, Andrea Hu, Jeffrey Hui, Jeremy Hurwitz, Michael Isard, Abe Ittycheriah, Matthew Jagielski, Wenhao Jia, Kathleen Kenealy, Maxim Krikun, Sneha Kudugunta, Chang
  Lan, Katherine Lee, Benjamin Lee, Eric Li, Music Li, Wei Li, YaGuang Li, Jian Li, Hyeontaek Lim, Hanzhao Lin, Zhongtao Liu, Frederick Liu, Marcello Maggioni, Aroma Mahendru, Joshua Maynez, Vedant Misra, Maysam Moussalem, Zachary Nado, John Nham, Eric Ni, Andrew Nystrom, Alicia Parrish, Marie Pellat, Martin Polacek, Alex Polozov, Reiner Pope, Siyuan Qiao, Emily Reif, Bryan Richter, Parker Riley, Alex~Castro Ros, Aurko Roy, Brennan Saeta, Rajkumar Samuel, Renee Shelby, Ambrose Slone, Daniel Smilkov, David~R. So, Daniel Sohn, Simon Tokumine, Dasha Valter, Vijay Vasudevan, Kiran Vodrahalli, Xuezhi Wang, Pidong Wang, Zirui Wang, Tao Wang, John Wieting, Yuhuai Wu, Kelvin Xu, Yunhan Xu, Linting Xue, Pengcheng Yin, Jiahui Yu, Qiao Zhang, Steven Zheng, Ce~Zheng, Weikang Zhou, Denny Zhou, Slav Petrov, and Yonghui Wu. 2023.
\newblock \href {http://arxiv.org/abs/2305.10403} {Palm 2 technical report}.

\bibitem[{Baevski and Auli(2019)}]{baevski2018adaptive}
Alexei Baevski and Michael Auli. 2019.
\newblock \href {https://openreview.net/forum?id=ByxZX20qFQ} {Adaptive input representations for neural language modeling}.
\newblock In \emph{International Conference on Learning Representations}.

\bibitem[{Bau et~al.(2020)Bau, Liu, Wang, Zhu, and Torralba}]{10.1007/978-3-030-58452-8_21}
David Bau, Steven Liu, Tongzhou Wang, Jun-Yan Zhu, and Antonio Torralba. 2020.
\newblock Rewriting a deep generative model.
\newblock In \emph{Computer Vision -- ECCV 2020}, pages 351--369, Cham. Springer International Publishing.

\bibitem[{Bhardwaj et~al.(2022)Bhardwaj, Polovets, and Sunkara}]{bhardwaj2022adaptation}
Rishabh Bhardwaj, George Polovets, and Monica Sunkara. 2022.
\newblock \href {http://arxiv.org/abs/2211.07828} {Adaptation approaches for nearest neighbor language models}.

\bibitem[{Biesialska et~al.(2020)Biesialska, Biesialska, and Costa-juss{\`a}}]{biesialska-etal-2020-continual}
Magdalena Biesialska, Katarzyna Biesialska, and Marta~R. Costa-juss{\`a}. 2020.
\newblock \href {https://doi.org/10.18653/v1/2020.coling-main.574} {Continual lifelong learning in natural language processing: A survey}.
\newblock In \emph{Proceedings of the 28th International Conference on Computational Linguistics}, pages 6523--6541, Barcelona, Spain (Online). International Committee on Computational Linguistics.

\bibitem[{Borgeaud et~al.(2022)Borgeaud, Mensch, Hoffmann, Cai, Rutherford, Millican, van~den Driessche, Lespiau, Damoc, Clark, de~Las~Casas, Guy, Menick, Ring, Hennigan, Huang, Maggiore, Jones, Cassirer, Brock, Paganini, Irving, Vinyals, Osindero, Simonyan, Rae, Elsen, and Sifre}]{borgeaud2022improving}
Sebastian Borgeaud, Arthur Mensch, Jordan Hoffmann, Trevor Cai, Eliza Rutherford, Katie Millican, George van~den Driessche, Jean-Baptiste Lespiau, Bogdan Damoc, Aidan Clark, Diego de~Las~Casas, Aurelia Guy, Jacob Menick, Roman Ring, Tom Hennigan, Saffron Huang, Loren Maggiore, Chris Jones, Albin Cassirer, Andy Brock, Michela Paganini, Geoffrey Irving, Oriol Vinyals, Simon Osindero, Karen Simonyan, Jack~W. Rae, Erich Elsen, and Laurent Sifre. 2022.
\newblock \href {http://arxiv.org/abs/2112.04426} {Improving language models by retrieving from trillions of tokens}.

\bibitem[{Brown et~al.(2023)Brown, Godfrey, Nizinski, Tu, and Kvinge}]{brown2023edit}
Davis Brown, Charles Godfrey, Cody Nizinski, Jonathan Tu, and Henry Kvinge. 2023.
\newblock \href {http://arxiv.org/abs/2303.00046} {Edit at your own risk: evaluating the robustness of edited models to distribution shifts}.

\bibitem[{Brown et~al.(2020)Brown, Mann, Ryder, Subbiah, Kaplan, Dhariwal, Neelakantan, Shyam, Sastry, Askell, Agarwal, Herbert-Voss, Krueger, Henighan, Child, Ramesh, Ziegler, Wu, Winter, Hesse, Chen, Sigler, Litwin, Gray, Chess, Clark, Berner, McCandlish, Radford, Sutskever, and Amodei}]{NEURIPS2020_1457c0d6}
Tom Brown, Benjamin Mann, Nick Ryder, Melanie Subbiah, Jared~D Kaplan, Prafulla Dhariwal, Arvind Neelakantan, Pranav Shyam, Girish Sastry, Amanda Askell, Sandhini Agarwal, Ariel Herbert-Voss, Gretchen Krueger, Tom Henighan, Rewon Child, Aditya Ramesh, Daniel Ziegler, Jeffrey Wu, Clemens Winter, Chris Hesse, Mark Chen, Eric Sigler, Mateusz Litwin, Scott Gray, Benjamin Chess, Jack Clark, Christopher Berner, Sam McCandlish, Alec Radford, Ilya Sutskever, and Dario Amodei. 2020.
\newblock \href {https://proceedings.neurips.cc/paper_files/paper/2020/file/1457c0d6bfcb4967418bfb8ac142f64a-Paper.pdf} {Language models are few-shot learners}.
\newblock In \emph{Advances in Neural Information Processing Systems}, volume~33, pages 1877--1901. Curran Associates, Inc.

\bibitem[{Bubeck et~al.(2023)Bubeck, Chandrasekaran, Eldan, Gehrke, Horvitz, Kamar, Lee, Lee, Li, Lundberg, Nori, Palangi, Ribeiro, and Zhang}]{bubeck2023sparks}
Sébastien Bubeck, Varun Chandrasekaran, Ronen Eldan, Johannes Gehrke, Eric Horvitz, Ece Kamar, Peter Lee, Yin~Tat Lee, Yuanzhi Li, Scott Lundberg, Harsha Nori, Hamid Palangi, Marco~Tulio Ribeiro, and Yi~Zhang. 2023.
\newblock \href {http://arxiv.org/abs/2303.12712} {Sparks of artificial general intelligence: Early experiments with gpt-4}.

\bibitem[{Cao et~al.(2023)Cao, Lin, Han, and Sun}]{cao2023life}
Boxi Cao, Hongyu Lin, Xianpei Han, and Le~Sun. 2023.
\newblock \href {http://arxiv.org/abs/2303.07616} {The life cycle of knowledge in big language models: A survey}.

\bibitem[{Carlini et~al.(2023)Carlini, Ippolito, Jagielski, Lee, Tramer, and Zhang}]{carlini2023quantifying}
Nicholas Carlini, Daphne Ippolito, Matthew Jagielski, Katherine Lee, Florian Tramer, and Chiyuan Zhang. 2023.
\newblock \href {https://openreview.net/forum?id=TatRHT_1cK} {Quantifying memorization across neural language models}.
\newblock In \emph{The Eleventh International Conference on Learning Representations}.

\bibitem[{Chase(2022)}]{langchain2022}
Harrison Chase. 2022.
\newblock \href {https://github.com/hwchase17/langchain} {Langchain}.

\bibitem[{Chen et~al.(2022)Chen, Zhang, and Choi}]{chen-etal-2022-rich}
Hung-Ting Chen, Michael Zhang, and Eunsol Choi. 2022.
\newblock \href {https://aclanthology.org/2022.emnlp-main.146} {Rich knowledge sources bring complex knowledge conflicts: Recalibrating models to reflect conflicting evidence}.
\newblock In \emph{Proceedings of the 2022 Conference on Empirical Methods in Natural Language Processing}, pages 2292--2307, Abu Dhabi, United Arab Emirates. Association for Computational Linguistics.

\bibitem[{Chen et~al.(2020)Chen, Hou, Cui, Che, Liu, and Yu}]{chen-etal-2020-recall}
Sanyuan Chen, Yutai Hou, Yiming Cui, Wanxiang Che, Ting Liu, and Xiangzhan Yu. 2020.
\newblock \href {https://doi.org/10.18653/v1/2020.emnlp-main.634} {Recall and learn: Fine-tuning deep pretrained language models with less forgetting}.
\newblock In \emph{Proceedings of the 2020 Conference on Empirical Methods in Natural Language Processing (EMNLP)}, pages 7870--7881, Online. Association for Computational Linguistics.

\bibitem[{Chen et~al.(2023{\natexlab{a}})Chen, Zhou, Du, Huang, Laudon, Chen, and Cui}]{pmlr-v202-chen23aq}
Wuyang Chen, Yanqi Zhou, Nan Du, Yanping Huang, James Laudon, Zhifeng Chen, and Claire Cui. 2023{\natexlab{a}}.
\newblock \href {https://proceedings.mlr.press/v202/chen23aq.html} {Lifelong language pretraining with distribution-specialized experts}.
\newblock In \emph{Proceedings of the 40th International Conference on Machine Learning}, volume 202 of \emph{Proceedings of Machine Learning Research}, pages 5383--5395. PMLR.

\bibitem[{Chen et~al.(2023{\natexlab{b}})Chen, Cao, Chen, Liu, and Zhao}]{chen2023journey}
Yuheng Chen, Pengfei Cao, Yubo Chen, Kang Liu, and Jun Zhao. 2023{\natexlab{b}}.
\newblock \href {http://arxiv.org/abs/2308.13198} {Journey to the center of the knowledge neurons: Discoveries of language-independent knowledge neurons and degenerate knowledge neurons}.

\bibitem[{Chen et~al.(2023{\natexlab{c}})Chen, Weiss, Mitchell, Celikyilmaz, and Bosselut}]{chen2023reckoning}
Zeming Chen, Gail Weiss, Eric Mitchell, Asli Celikyilmaz, and Antoine Bosselut. 2023{\natexlab{c}}.
\newblock \href {http://arxiv.org/abs/2305.06349} {Reckoning: Reasoning through dynamic knowledge encoding}.

\bibitem[{Chen et~al.(2023{\natexlab{d}})Chen, Zhou, Zhang, Gong, Zhao, and Wen}]{chen2023chatcot}
Zhipeng Chen, Kun Zhou, Beichen Zhang, Zheng Gong, Wayne~Xin Zhao, and Ji-Rong Wen. 2023{\natexlab{d}}.
\newblock \href {http://arxiv.org/abs/2305.14323} {Chatcot: Tool-augmented chain-of-thought reasoning on chat-based large language models}.

\bibitem[{Cheng et~al.(2023)Cheng, Lin, Chen, Zhao, and Yan}]{cheng2023decouple}
Xin Cheng, Yankai Lin, Xiuying Chen, Dongyan Zhao, and Rui Yan. 2023.
\newblock \href {http://arxiv.org/abs/2305.11564} {Decouple knowledge from paramters for plug-and-play language modeling}.

\bibitem[{Chowdhery et~al.(2022)Chowdhery, Narang, Devlin, Bosma, Mishra, Roberts, Barham, Chung, Sutton, Gehrmann, Schuh, Shi, Tsvyashchenko, Maynez, Rao, Barnes, Tay, Shazeer, Prabhakaran, Reif, Du, Hutchinson, Pope, Bradbury, Austin, Isard, Gur-Ari, Yin, Duke, Levskaya, Ghemawat, Dev, Michalewski, Garcia, Misra, Robinson, Fedus, Zhou, Ippolito, Luan, Lim, Zoph, Spiridonov, Sepassi, Dohan, Agrawal, Omernick, Dai, Pillai, Pellat, Lewkowycz, Moreira, Child, Polozov, Lee, Zhou, Wang, Saeta, Diaz, Firat, Catasta, Wei, Meier-Hellstern, Eck, Dean, Petrov, and Fiedel}]{chowdhery2022palm}
Aakanksha Chowdhery, Sharan Narang, Jacob Devlin, Maarten Bosma, Gaurav Mishra, Adam Roberts, Paul Barham, Hyung~Won Chung, Charles Sutton, Sebastian Gehrmann, Parker Schuh, Kensen Shi, Sasha Tsvyashchenko, Joshua Maynez, Abhishek Rao, Parker Barnes, Yi~Tay, Noam Shazeer, Vinodkumar Prabhakaran, Emily Reif, Nan Du, Ben Hutchinson, Reiner Pope, James Bradbury, Jacob Austin, Michael Isard, Guy Gur-Ari, Pengcheng Yin, Toju Duke, Anselm Levskaya, Sanjay Ghemawat, Sunipa Dev, Henryk Michalewski, Xavier Garcia, Vedant Misra, Kevin Robinson, Liam Fedus, Denny Zhou, Daphne Ippolito, David Luan, Hyeontaek Lim, Barret Zoph, Alexander Spiridonov, Ryan Sepassi, David Dohan, Shivani Agrawal, Mark Omernick, Andrew~M. Dai, Thanumalayan~Sankaranarayana Pillai, Marie Pellat, Aitor Lewkowycz, Erica Moreira, Rewon Child, Oleksandr Polozov, Katherine Lee, Zongwei Zhou, Xuezhi Wang, Brennan Saeta, Mark Diaz, Orhan Firat, Michele Catasta, Jason Wei, Kathy Meier-Hellstern, Douglas Eck, Jeff Dean, Slav Petrov, and Noah Fiedel. 2022.
\newblock \href {http://arxiv.org/abs/2204.02311} {Palm: Scaling language modeling with pathways}.

\bibitem[{Cohen et~al.(2023)Cohen, Biran, Yoran, Globerson, and Geva}]{cohen2023evaluating}
Roi Cohen, Eden Biran, Ori Yoran, Amir Globerson, and Mor Geva. 2023.
\newblock \href {http://arxiv.org/abs/2307.12976} {Evaluating the ripple effects of knowledge editing in language models}.

\bibitem[{Dai et~al.(2022)Dai, Dong, Hao, Sui, Chang, and Wei}]{dai-etal-2022-knowledge}
Damai Dai, Li~Dong, Yaru Hao, Zhifang Sui, Baobao Chang, and Furu Wei. 2022.
\newblock \href {https://doi.org/10.18653/v1/2022.acl-long.581} {Knowledge neurons in pretrained transformers}.
\newblock In \emph{Proceedings of the 60th Annual Meeting of the Association for Computational Linguistics (Volume 1: Long Papers)}, pages 8493--8502, Dublin, Ireland. Association for Computational Linguistics.

\bibitem[{Dalvi~Mishra et~al.(2022)Dalvi~Mishra, Tafjord, and Clark}]{dalvi-mishra-etal-2022-towards}
Bhavana Dalvi~Mishra, Oyvind Tafjord, and Peter Clark. 2022.
\newblock \href {https://aclanthology.org/2022.emnlp-main.644} {Towards teachable reasoning systems: Using a dynamic memory of user feedback for continual system improvement}.
\newblock In \emph{Proceedings of the 2022 Conference on Empirical Methods in Natural Language Processing}, pages 9465--9480, Abu Dhabi, United Arab Emirates. Association for Computational Linguistics.

\bibitem[{De~Cao et~al.(2021)De~Cao, Aziz, and Titov}]{de-cao-etal-2021-editing}
Nicola De~Cao, Wilker Aziz, and Ivan Titov. 2021.
\newblock \href {https://doi.org/10.18653/v1/2021.emnlp-main.522} {Editing factual knowledge in language models}.
\newblock In \emph{Proceedings of the 2021 Conference on Empirical Methods in Natural Language Processing}, pages 6491--6506, Online and Punta Cana, Dominican Republic. Association for Computational Linguistics.

\bibitem[{Devlin et~al.(2019)Devlin, Chang, Lee, and Toutanova}]{devlin-etal-2019-bert}
Jacob Devlin, Ming-Wei Chang, Kenton Lee, and Kristina Toutanova. 2019.
\newblock \href {https://doi.org/10.18653/v1/N19-1423} {{BERT}: Pre-training of deep bidirectional transformers for language understanding}.
\newblock In \emph{Proceedings of the 2019 Conference of the North {A}merican Chapter of the Association for Computational Linguistics: Human Language Technologies, Volume 1 (Long and Short Papers)}, pages 4171--4186, Minneapolis, Minnesota. Association for Computational Linguistics.

\bibitem[{Dhingra et~al.(2022)Dhingra, Cole, Eisenschlos, Gillick, Eisenstein, and Cohen}]{dhingra-etal-2022-time}
Bhuwan Dhingra, Jeremy~R. Cole, Julian~Martin Eisenschlos, Daniel Gillick, Jacob Eisenstein, and William~W. Cohen. 2022.
\newblock \href {https://doi.org/10.1162/tacl_a_00459} {Time-aware language models as temporal knowledge bases}.
\newblock \emph{Transactions of the Association for Computational Linguistics}, 10:257--273.

\bibitem[{Ding et~al.(2022)Ding, Qin, Yang, Wei, Yang, Su, Hu, Chen, Chan, Chen, Yi, Zhao, Wang, Liu, Zheng, Chen, Liu, Tang, Li, and Sun}]{ding2022delta}
Ning Ding, Yujia Qin, Guang Yang, Fuchao Wei, Zonghan Yang, Yusheng Su, Shengding Hu, Yulin Chen, Chi-Min Chan, Weize Chen, Jing Yi, Weilin Zhao, Xiaozhi Wang, Zhiyuan Liu, Hai-Tao Zheng, Jianfei Chen, Yang Liu, Jie Tang, Juanzi Li, and Maosong Sun. 2022.
\newblock \href {http://arxiv.org/abs/2203.06904} {Delta tuning: A comprehensive study of parameter efficient methods for pre-trained language models}.

\bibitem[{Dong et~al.(2022)Dong, Dai, Song, Xu, Sui, and Li}]{dong-etal-2022-calibrating}
Qingxiu Dong, Damai Dai, Yifan Song, Jingjing Xu, Zhifang Sui, and Lei Li. 2022.
\newblock \href {https://aclanthology.org/2022.findings-emnlp.438} {Calibrating factual knowledge in pretrained language models}.
\newblock In \emph{Findings of the Association for Computational Linguistics: EMNLP 2022}, pages 5937--5947, Abu Dhabi, United Arab Emirates. Association for Computational Linguistics.

\bibitem[{Drozdov et~al.(2022)Drozdov, Wang, Rahimi, McCallum, Zamani, and Iyyer}]{drozdov-etal-2022-cant}
Andrew Drozdov, Shufan Wang, Razieh Rahimi, Andrew McCallum, Hamed Zamani, and Mohit Iyyer. 2022.
\newblock \href {https://aclanthology.org/2022.findings-emnlp.218} {You can{'}t pick your neighbors, or can you? when and how to rely on retrieval in the k{NN}-{LM}}.
\newblock In \emph{Findings of the Association for Computational Linguistics: EMNLP 2022}, pages 2997--3007, Abu Dhabi, United Arab Emirates. Association for Computational Linguistics.

\bibitem[{Feng et~al.(2023)Feng, Zhang, and Fei}]{feng2023knowledge}
Chao Feng, Xinyu Zhang, and Zichu Fei. 2023.
\newblock \href {http://arxiv.org/abs/2309.03118} {Knowledge solver: Teaching llms to search for domain knowledge from knowledge graphs}.

\bibitem[{Finn et~al.(2017)Finn, Abbeel, and Levine}]{10.5555/3305381.3305498}
Chelsea Finn, Pieter Abbeel, and Sergey Levine. 2017.
\newblock Model-agnostic meta-learning for fast adaptation of deep networks.
\newblock In \emph{Proceedings of the 34th International Conference on Machine Learning - Volume 70}, ICML'17, page 1126–1135. JMLR.org.

\bibitem[{Gao et~al.(2023)Gao, Dai, Pasupat, Chen, Chaganty, Fan, Zhao, Lao, Lee, Juan, and Guu}]{gao2023rarr}
Luyu Gao, Zhuyun Dai, Panupong Pasupat, Anthony Chen, Arun~Tejasvi Chaganty, Yicheng Fan, Vincent~Y. Zhao, Ni~Lao, Hongrae Lee, Da-Cheng Juan, and Kelvin Guu. 2023.
\newblock \href {http://arxiv.org/abs/2210.08726} {Rarr: Researching and revising what language models say, using language models}.

\bibitem[{Geva et~al.(2023)Geva, Bastings, Filippova, and Globerson}]{geva2023dissecting}
Mor Geva, Jasmijn Bastings, Katja Filippova, and Amir Globerson. 2023.
\newblock \href {http://arxiv.org/abs/2304.14767} {Dissecting recall of factual associations in auto-regressive language models}.

\bibitem[{Geva et~al.(2021)Geva, Schuster, Berant, and Levy}]{geva-etal-2021-transformer}
Mor Geva, Roei Schuster, Jonathan Berant, and Omer Levy. 2021.
\newblock \href {https://doi.org/10.18653/v1/2021.emnlp-main.446} {Transformer feed-forward layers are key-value memories}.
\newblock In \emph{Proceedings of the 2021 Conference on Empirical Methods in Natural Language Processing}, pages 5484--5495, Online and Punta Cana, Dominican Republic. Association for Computational Linguistics.

\bibitem[{Google(2023)}]{Med_PaLM_2}
Google. 2023.
\newblock \href {https://cloud.google.com/blog/topics/healthcare-life-sciences/sharing-google-med-palm-2-medical-large-language-model} {Med-palm 2}.

\bibitem[{Gou et~al.(2023)Gou, Shao, Gong, Shen, Yang, Duan, and Chen}]{gou2023critic}
Zhibin Gou, Zhihong Shao, Yeyun Gong, Yelong Shen, Yujiu Yang, Nan Duan, and Weizhu Chen. 2023.
\newblock \href {http://arxiv.org/abs/2305.11738} {Critic: Large language models can self-correct with tool-interactive critiquing}.

\bibitem[{Gupta et~al.(2023{\natexlab{a}})Gupta, Mondal, Sheshadri, Zhao, Li, Wiegreffe, and Tandon}]{gupta2023editing}
Anshita Gupta, Debanjan Mondal, Akshay~Krishna Sheshadri, Wenlong Zhao, Xiang~Lorraine Li, Sarah Wiegreffe, and Niket Tandon. 2023{\natexlab{a}}.
\newblock \href {http://arxiv.org/abs/2305.14956} {Editing commonsense knowledge in gpt}.

\bibitem[{Gupta et~al.(2023{\natexlab{b}})Gupta, Thérien, Ibrahim, Richter, Anthony, Belilovsky, Rish, and Lesort}]{gupta2023continual}
Kshitij Gupta, Benjamin Thérien, Adam Ibrahim, Mats~L. Richter, Quentin Anthony, Eugene Belilovsky, Irina Rish, and Timothée Lesort. 2023{\natexlab{b}}.
\newblock \href {http://arxiv.org/abs/2308.04014} {Continual pre-training of large language models: How to (re)warm your model?}

\bibitem[{Gururangan et~al.(2022)Gururangan, Lewis, Holtzman, Smith, and Zettlemoyer}]{gururangan-etal-2022-demix}
Suchin Gururangan, Mike Lewis, Ari Holtzman, Noah~A. Smith, and Luke Zettlemoyer. 2022.
\newblock \href {https://doi.org/10.18653/v1/2022.naacl-main.407} {{DEM}ix layers: Disentangling domains for modular language modeling}.
\newblock In \emph{Proceedings of the 2022 Conference of the North American Chapter of the Association for Computational Linguistics: Human Language Technologies}, pages 5557--5576, Seattle, United States. Association for Computational Linguistics.

\bibitem[{Gururangan et~al.(2020)Gururangan, Marasovi{\'c}, Swayamdipta, Lo, Beltagy, Downey, and Smith}]{gururangan-etal-2020-dont}
Suchin Gururangan, Ana Marasovi{\'c}, Swabha Swayamdipta, Kyle Lo, Iz~Beltagy, Doug Downey, and Noah~A. Smith. 2020.
\newblock \href {https://doi.org/10.18653/v1/2020.acl-main.740} {Don{'}t stop pretraining: Adapt language models to domains and tasks}.
\newblock In \emph{Proceedings of the 58th Annual Meeting of the Association for Computational Linguistics}, pages 8342--8360, Online. Association for Computational Linguistics.

\bibitem[{Guu et~al.(2020)Guu, Lee, Tung, Pasupat, and Chang}]{guu2020retrieval}
Kelvin Guu, Kenton Lee, Zora Tung, Panupong Pasupat, and Mingwei Chang. 2020.
\newblock Retrieval augmented language model pre-training.
\newblock In \emph{International conference on machine learning}, pages 3929--3938. PMLR.

\bibitem[{Ha et~al.(2017)Ha, Dai, and Le}]{ha2017hypernetworks}
David Ha, Andrew~M. Dai, and Quoc~V. Le. 2017.
\newblock \href {https://openreview.net/forum?id=rkpACe1lx} {Hypernetworks}.
\newblock In \emph{International Conference on Learning Representations}.

\bibitem[{Hartvigsen et~al.(2023)Hartvigsen, Sankaranarayanan, Palangi, Kim, and Ghassemi}]{hartvigsen2023aging}
Thomas Hartvigsen, Swami Sankaranarayanan, Hamid Palangi, Yoon Kim, and Marzyeh Ghassemi. 2023.
\newblock \href {http://arxiv.org/abs/2211.11031} {Aging with grace: Lifelong model editing with discrete key-value adaptors}.

\bibitem[{Hase et~al.(2023{\natexlab{a}})Hase, Bansal, Kim, and Ghandeharioun}]{hase2023does}
Peter Hase, Mohit Bansal, Been Kim, and Asma Ghandeharioun. 2023{\natexlab{a}}.
\newblock Does localization inform editing? surprising differences in causality-based localization vs. knowledge editing in language models.
\newblock \emph{arXiv preprint arXiv:2301.04213}.

\bibitem[{Hase et~al.(2023{\natexlab{b}})Hase, Diab, Celikyilmaz, Li, Kozareva, Stoyanov, Bansal, and Iyer}]{hase-etal-2023-methods}
Peter Hase, Mona Diab, Asli Celikyilmaz, Xian Li, Zornitsa Kozareva, Veselin Stoyanov, Mohit Bansal, and Srinivasan Iyer. 2023{\natexlab{b}}.
\newblock \href {https://aclanthology.org/2023.eacl-main.199} {Methods for measuring, updating, and visualizing factual beliefs in language models}.
\newblock In \emph{Proceedings of the 17th Conference of the European Chapter of the Association for Computational Linguistics}, pages 2714--2731, Dubrovnik, Croatia. Association for Computational Linguistics.

\bibitem[{He et~al.(2022)He, Zhang, and Roth}]{he2022rethinking}
Hangfeng He, Hongming Zhang, and Dan Roth. 2022.
\newblock \href {http://arxiv.org/abs/2301.00303} {Rethinking with retrieval: Faithful large language model inference}.

\bibitem[{He et~al.(2021{\natexlab{a}})He, Neubig, and Berg-Kirkpatrick}]{he-etal-2021-efficient}
Junxian He, Graham Neubig, and Taylor Berg-Kirkpatrick. 2021{\natexlab{a}}.
\newblock \href {https://doi.org/10.18653/v1/2021.emnlp-main.461} {Efficient nearest neighbor language models}.
\newblock In \emph{Proceedings of the 2021 Conference on Empirical Methods in Natural Language Processing}, pages 5703--5714, Online and Punta Cana, Dominican Republic. Association for Computational Linguistics.

\bibitem[{He et~al.(2021{\natexlab{b}})He, Liu, Cho, Ott, Liu, Glass, and Peng}]{he-etal-2021-analyzing}
Tianxing He, Jun Liu, Kyunghyun Cho, Myle Ott, Bing Liu, James Glass, and Fuchun Peng. 2021{\natexlab{b}}.
\newblock \href {https://doi.org/10.18653/v1/2021.eacl-main.95} {Analyzing the forgetting problem in pretrain-finetuning of open-domain dialogue response models}.
\newblock In \emph{Proceedings of the 16th Conference of the European Chapter of the Association for Computational Linguistics: Main Volume}, pages 1121--1133, Online. Association for Computational Linguistics.

\bibitem[{Hernandez et~al.(2023)Hernandez, Li, and Andreas}]{hernandez2023inspecting}
Evan Hernandez, Belinda~Z. Li, and Jacob Andreas. 2023.
\newblock \href {http://arxiv.org/abs/2304.00740} {Inspecting and editing knowledge representations in language models}.

\bibitem[{Hoelscher-Obermaier et~al.(2023)Hoelscher-Obermaier, Persson, Kran, Konstas, and Barez}]{hoelscherobermaier2023detecting}
Jason Hoelscher-Obermaier, Julia Persson, Esben Kran, Ioannis Konstas, and Fazl Barez. 2023.
\newblock \href {http://arxiv.org/abs/2305.17553} {Detecting edit failures in large language models: An improved specificity benchmark}.

\bibitem[{Houlsby et~al.(2019)Houlsby, Giurgiu, Jastrzebski, Morrone, De~Laroussilhe, Gesmundo, Attariyan, and Gelly}]{houlsby2019parameter}
Neil Houlsby, Andrei Giurgiu, Stanislaw Jastrzebski, Bruna Morrone, Quentin De~Laroussilhe, Andrea Gesmundo, Mona Attariyan, and Sylvain Gelly. 2019.
\newblock Parameter-efficient transfer learning for nlp.
\newblock In \emph{International Conference on Machine Learning}, pages 2790--2799. PMLR.

\bibitem[{Hu et~al.(2022)Hu, yelong shen, Wallis, Allen-Zhu, Li, Wang, Wang, and Chen}]{hu2022lora}
Edward~J Hu, yelong shen, Phillip Wallis, Zeyuan Allen-Zhu, Yuanzhi Li, Shean Wang, Lu~Wang, and Weizhu Chen. 2022.
\newblock \href {https://openreview.net/forum?id=nZeVKeeFYf9} {Lo{RA}: Low-rank adaptation of large language models}.
\newblock In \emph{International Conference on Learning Representations}.

\bibitem[{Hu et~al.(2023)Hu, Mitchell, Manning, and Finn}]{hu2023metalearning}
Nathan Hu, Eric Mitchell, Christopher~D. Manning, and Chelsea Finn. 2023.
\newblock \href {http://arxiv.org/abs/2305.15076} {Meta-learning online adaptation of language models}.

\bibitem[{Huang et~al.(2023)Huang, Shen, Zhang, Zhou, Rong, and Xiong}]{huang2023transformerpatcher}
Zeyu Huang, Yikang Shen, Xiaofeng Zhang, Jie Zhou, Wenge Rong, and Zhang Xiong. 2023.
\newblock \href {https://openreview.net/forum?id=4oYUGeGBPm} {Transformer-patcher: One mistake worth one neuron}.
\newblock In \emph{The Eleventh International Conference on Learning Representations}.

\bibitem[{Inaba et~al.(2023)Inaba, Kiyomaru, Cheng, and Kurohashi}]{inaba2023multitoolcot}
Tatsuro Inaba, Hirokazu Kiyomaru, Fei Cheng, and Sadao Kurohashi. 2023.
\newblock \href {http://arxiv.org/abs/2305.16896} {Multitool-cot: Gpt-3 can use multiple external tools with chain of thought prompting}.

\bibitem[{Izacard et~al.(2022)Izacard, Lewis, Lomeli, Hosseini, Petroni, Schick, Dwivedi-Yu, Joulin, Riedel, and Grave}]{izacard2022atlas}
Gautier Izacard, Patrick Lewis, Maria Lomeli, Lucas Hosseini, Fabio Petroni, Timo Schick, Jane Dwivedi-Yu, Armand Joulin, Sebastian Riedel, and Edouard Grave. 2022.
\newblock \href {http://arxiv.org/abs/2208.03299} {Atlas: Few-shot learning with retrieval augmented language models}.

\bibitem[{Jang et~al.(2022{\natexlab{a}})Jang, Ye, Lee, Yang, Shin, Han, Kim, and Seo}]{jang-etal-2022-temporalwiki}
Joel Jang, Seonghyeon Ye, Changho Lee, Sohee Yang, Joongbo Shin, Janghoon Han, Gyeonghun Kim, and Minjoon Seo. 2022{\natexlab{a}}.
\newblock \href {https://aclanthology.org/2022.emnlp-main.418} {{T}emporal{W}iki: A lifelong benchmark for training and evaluating ever-evolving language models}.
\newblock In \emph{Proceedings of the 2022 Conference on Empirical Methods in Natural Language Processing}, pages 6237--6250, Abu Dhabi, United Arab Emirates. Association for Computational Linguistics.

\bibitem[{Jang et~al.(2022{\natexlab{b}})Jang, Ye, Yang, Shin, Han, KIM, Choi, and Seo}]{jang2022towards}
Joel Jang, Seonghyeon Ye, Sohee Yang, Joongbo Shin, Janghoon Han, Gyeonghun KIM, Stanley~Jungkyu Choi, and Minjoon Seo. 2022{\natexlab{b}}.
\newblock \href {https://openreview.net/forum?id=vfsRB5MImo9} {Towards continual knowledge learning of language models}.
\newblock In \emph{International Conference on Learning Representations}.

\bibitem[{Ji et~al.(2022)Ji, Pan, Cambria, Marttinen, and Yu}]{Ji_2022}
Shaoxiong Ji, Shirui Pan, Erik Cambria, Pekka Marttinen, and Philip~S. Yu. 2022.
\newblock \href {https://doi.org/10.1109/tnnls.2021.3070843} {A survey on knowledge graphs: Representation, acquisition, and applications}.
\newblock \emph{{IEEE} Transactions on Neural Networks and Learning Systems}, 33(2):494--514.

\bibitem[{Ji et~al.(2023)Ji, Lee, Frieske, Yu, Su, Xu, Ishii, Bang, Madotto, and Fung}]{10.1145/3571730}
Ziwei Ji, Nayeon Lee, Rita Frieske, Tiezheng Yu, Dan Su, Yan Xu, Etsuko Ishii, Ye~Jin Bang, Andrea Madotto, and Pascale Fung. 2023.
\newblock \href {https://doi.org/10.1145/3571730} {Survey of hallucination in natural language generation}.
\newblock \emph{ACM Comput. Surv.}, 55(12).

\bibitem[{Jiang et~al.(2022)Jiang, Gao, Wang, Araki, Ding, Callan, and Neubig}]{jiang-etal-2022-retrieval}
Zhengbao Jiang, Luyu Gao, Zhiruo Wang, Jun Araki, Haibo Ding, Jamie Callan, and Graham Neubig. 2022.
\newblock \href {https://aclanthology.org/2022.emnlp-main.149} {Retrieval as attention: End-to-end learning of retrieval and reading within a single transformer}.
\newblock In \emph{Proceedings of the 2022 Conference on Empirical Methods in Natural Language Processing}, pages 2336--2349, Abu Dhabi, United Arab Emirates. Association for Computational Linguistics.

\bibitem[{Jiang et~al.(2020)Jiang, Xu, Araki, and Neubig}]{jiang-etal-2020-know}
Zhengbao Jiang, Frank~F. Xu, Jun Araki, and Graham Neubig. 2020.
\newblock \href {https://doi.org/10.1162/tacl_a_00324} {How can we know what language models know?}
\newblock \emph{Transactions of the Association for Computational Linguistics}, 8:423--438.

\bibitem[{Jiang et~al.(2023)Jiang, Xu, Gao, Sun, Liu, Dwivedi-Yu, Yang, Callan, and Neubig}]{jiang2023active}
Zhengbao Jiang, Frank~F. Xu, Luyu Gao, Zhiqing Sun, Qian Liu, Jane Dwivedi-Yu, Yiming Yang, Jamie Callan, and Graham Neubig. 2023.
\newblock \href {http://arxiv.org/abs/2305.06983} {Active retrieval augmented generation}.

\bibitem[{Jin et~al.(2022)Jin, Zhang, Zhu, Xiao, Li, Wei, Arnold, and Ren}]{jin-etal-2022-lifelong-pretraining}
Xisen Jin, Dejiao Zhang, Henghui Zhu, Wei Xiao, Shang-Wen Li, Xiaokai Wei, Andrew Arnold, and Xiang Ren. 2022.
\newblock \href {https://doi.org/10.18653/v1/2022.naacl-main.351} {Lifelong pretraining: Continually adapting language models to emerging corpora}.
\newblock In \emph{Proceedings of the 2022 Conference of the North American Chapter of the Association for Computational Linguistics: Human Language Technologies}, pages 4764--4780, Seattle, United States. Association for Computational Linguistics.

\bibitem[{Ju and Zhang(2023)}]{ju2023klob}
Yiming Ju and Zheng Zhang. 2023.
\newblock \href {http://arxiv.org/abs/2309.16535} {Klob: a benchmark for assessing knowledge locating methods in language models}.

\bibitem[{Kamalloo et~al.(2023)Kamalloo, Dziri, Clarke, and Rafiei}]{kamalloo2023evaluating}
Ehsan Kamalloo, Nouha Dziri, Charles L.~A. Clarke, and Davood Rafiei. 2023.
\newblock \href {http://arxiv.org/abs/2305.06984} {Evaluating open-domain question answering in the era of large language models}.

\bibitem[{Kang et~al.(2023)Kang, Laroche, Yuan, Trischler, Liu, and Fu}]{kang2023think}
Jikun Kang, Romain Laroche, Xindi Yuan, Adam Trischler, Xue Liu, and Jie Fu. 2023.
\newblock \href {http://arxiv.org/abs/2305.16338} {Think before you act: Decision transformers with internal working memory}.

\bibitem[{Karpukhin et~al.(2020)Karpukhin, Oguz, Min, Lewis, Wu, Edunov, Chen, and Yih}]{karpukhin-etal-2020-dense}
Vladimir Karpukhin, Barlas Oguz, Sewon Min, Patrick Lewis, Ledell Wu, Sergey Edunov, Danqi Chen, and Wen-tau Yih. 2020.
\newblock \href {https://doi.org/10.18653/v1/2020.emnlp-main.550} {Dense passage retrieval for open-domain question answering}.
\newblock In \emph{Proceedings of the 2020 Conference on Empirical Methods in Natural Language Processing (EMNLP)}, pages 6769--6781, Online. Association for Computational Linguistics.

\bibitem[{Kasai et~al.(2022)Kasai, Sakaguchi, Takahashi, Bras, Asai, Yu, Radev, Smith, Choi, and Inui}]{kasai2022realtime}
Jungo Kasai, Keisuke Sakaguchi, Yoichi Takahashi, Ronan~Le Bras, Akari Asai, Xinyan Yu, Dragomir Radev, Noah~A. Smith, Yejin Choi, and Kentaro Inui. 2022.
\newblock \href {http://arxiv.org/abs/2207.13332} {Realtime qa: What's the answer right now?}

\bibitem[{Kassner et~al.(2021)Kassner, Tafjord, Sch{\"u}tze, and Clark}]{kassner-etal-2021-beliefbank}
Nora Kassner, Oyvind Tafjord, Hinrich Sch{\"u}tze, and Peter Clark. 2021.
\newblock \href {https://doi.org/10.18653/v1/2021.emnlp-main.697} {{B}elief{B}ank: Adding memory to a pre-trained language model for a systematic notion of belief}.
\newblock In \emph{Proceedings of the 2021 Conference on Empirical Methods in Natural Language Processing}, pages 8849--8861, Online and Punta Cana, Dominican Republic. Association for Computational Linguistics.

\bibitem[{Kaur et~al.(2022)Kaur, Bhatia, Aggarwal, Bansal, and Krishnamurthy}]{kaur-etal-2022-lm}
Jivat Kaur, Sumit Bhatia, Milan Aggarwal, Rachit Bansal, and Balaji Krishnamurthy. 2022.
\newblock \href {https://doi.org/10.18653/v1/2022.findings-naacl.57} {{LM}-{CORE}: Language models with contextually relevant external knowledge}.
\newblock In \emph{Findings of the Association for Computational Linguistics: NAACL 2022}, pages 750--769, Seattle, United States. Association for Computational Linguistics.

\bibitem[{Ke et~al.(2022)Ke, Lin, Shao, Xu, Shu, and Liu}]{ke-etal-2022-continual}
Zixuan Ke, Haowei Lin, Yijia Shao, Hu~Xu, Lei Shu, and Bing Liu. 2022.
\newblock \href {https://aclanthology.org/2022.emnlp-main.695} {Continual training of language models for few-shot learning}.
\newblock In \emph{Proceedings of the 2022 Conference on Empirical Methods in Natural Language Processing}, pages 10205--10216, Abu Dhabi, United Arab Emirates. Association for Computational Linguistics.

\bibitem[{Ke and Liu(2023)}]{ke2023continua_surveyl}
Zixuan Ke and Bing Liu. 2023.
\newblock \href {http://arxiv.org/abs/2211.12701} {Continual learning of natural language processing tasks: A survey}.

\bibitem[{Ke et~al.(2023)Ke, Shao, Lin, Konishi, Kim, and Liu}]{ke2023continual}
Zixuan Ke, Yijia Shao, Haowei Lin, Tatsuya Konishi, Gyuhak Kim, and Bing Liu. 2023.
\newblock \href {https://openreview.net/forum?id=m_GDIItaI3o} {Continual pre-training of language models}.
\newblock In \emph{The Eleventh International Conference on Learning Representations}.

\bibitem[{Khandelwal et~al.(2020)Khandelwal, Levy, Jurafsky, Zettlemoyer, and Lewis}]{Khandelwal2020Generalization}
Urvashi Khandelwal, Omer Levy, Dan Jurafsky, Luke Zettlemoyer, and Mike Lewis. 2020.
\newblock \href {https://openreview.net/forum?id=HklBjCEKvH} {Generalization through memorization: Nearest neighbor language models}.
\newblock In \emph{International Conference on Learning Representations}.

\bibitem[{Khattab et~al.(2023)Khattab, Santhanam, Li, Hall, Liang, Potts, and Zaharia}]{khattab2023demonstratesearchpredict}
Omar Khattab, Keshav Santhanam, Xiang~Lisa Li, David Hall, Percy Liang, Christopher Potts, and Matei Zaharia. 2023.
\newblock \href {http://arxiv.org/abs/2212.14024} {Demonstrate-search-predict: Composing retrieval and language models for knowledge-intensive nlp}.

\bibitem[{Khot et~al.(2023)Khot, Trivedi, Finlayson, Fu, Richardson, Clark, and Sabharwal}]{khot2023decomposed}
Tushar Khot, Harsh Trivedi, Matthew Finlayson, Yao Fu, Kyle Richardson, Peter Clark, and Ashish Sabharwal. 2023.
\newblock \href {https://openreview.net/forum?id=_nGgzQjzaRy} {Decomposed prompting: A modular approach for solving complex tasks}.
\newblock In \emph{The Eleventh International Conference on Learning Representations}.

\bibitem[{Kirkpatrick et~al.(2017)Kirkpatrick, Pascanu, Rabinowitz, Veness, Desjardins, Rusu, Milan, Quan, Ramalho, Grabska-Barwinska, Hassabis, Clopath, Kumaran, and Hadsell}]{doi:10.1073/pnas.1611835114}
James Kirkpatrick, Razvan Pascanu, Neil Rabinowitz, Joel Veness, Guillaume Desjardins, Andrei~A. Rusu, Kieran Milan, John Quan, Tiago Ramalho, Agnieszka Grabska-Barwinska, Demis Hassabis, Claudia Clopath, Dharshan Kumaran, and Raia Hadsell. 2017.
\newblock \href {https://doi.org/10.1073/pnas.1611835114} {Overcoming catastrophic forgetting in neural networks}.
\newblock \emph{Proceedings of the National Academy of Sciences}, 114(13):3521--3526.

\bibitem[{Komeili et~al.(2022)Komeili, Shuster, and Weston}]{komeili-etal-2022-internet}
Mojtaba Komeili, Kurt Shuster, and Jason Weston. 2022.
\newblock \href {https://doi.org/10.18653/v1/2022.acl-long.579} {{I}nternet-augmented dialogue generation}.
\newblock In \emph{Proceedings of the 60th Annual Meeting of the Association for Computational Linguistics (Volume 1: Long Papers)}, pages 8460--8478, Dublin, Ireland. Association for Computational Linguistics.

\bibitem[{Lazaridou et~al.(2022)Lazaridou, Gribovskaya, Stokowiec, and Grigorev}]{lazaridou2022internetaugmented}
Angeliki Lazaridou, Elena Gribovskaya, Wojciech Stokowiec, and Nikolai Grigorev. 2022.
\newblock \href {http://arxiv.org/abs/2203.05115} {Internet-augmented language models through few-shot prompting for open-domain question answering}.

\bibitem[{Lazaridou et~al.(2021)Lazaridou, Kuncoro, Gribovskaya, Agrawal, Liska, Terzi, Gimenez, de~Masson~d'Autume, Ko{\v{c}}isk{\'y}, Ruder, Yogatama, Cao, Young, and Blunsom}]{lazaridou2021mind}
Angeliki Lazaridou, Adhiguna Kuncoro, Elena Gribovskaya, Devang Agrawal, Adam Liska, Tayfun Terzi, Mai Gimenez, Cyprien de~Masson~d'Autume, Tom{\'a}{\v{s}} Ko{\v{c}}isk{\'y}, Sebastian Ruder, Dani Yogatama, Kris Cao, Susannah Young, and Phil Blunsom. 2021.
\newblock \href {https://openreview.net/forum?id=73OmmrCfSyy} {Mind the gap: Assessing temporal generalization in neural language models}.
\newblock In \emph{Advances in Neural Information Processing Systems}.

\bibitem[{Lee et~al.(2022{\natexlab{a}})Lee, Han, Hwang, Lee, Park, and Lee}]{lee-etal-2022-plug}
Kyungjae Lee, Wookje Han, Seung-won Hwang, Hwaran Lee, Joonsuk Park, and Sang-Woo Lee. 2022{\natexlab{a}}.
\newblock \href {https://doi.org/10.18653/v1/2022.findings-acl.37} {Plug-and-play adaptation for continuously-updated {QA}}.
\newblock In \emph{Findings of the Association for Computational Linguistics: ACL 2022}, pages 438--447, Dublin, Ireland. Association for Computational Linguistics.

\bibitem[{Lee et~al.(2022{\natexlab{b}})Lee, Ping, Xu, Patwary, Fung, Shoeybi, and Catanzaro}]{NEURIPS2022_df438caa}
Nayeon Lee, Wei Ping, Peng Xu, Mostofa Patwary, Pascale~N Fung, Mohammad Shoeybi, and Bryan Catanzaro. 2022{\natexlab{b}}.
\newblock \href {https://proceedings.neurips.cc/paper_files/paper/2022/file/df438caa36714f69277daa92d608dd63-Paper-Conference.pdf} {Factuality enhanced language models for open-ended text generation}.
\newblock In \emph{Advances in Neural Information Processing Systems}, volume~35, pages 34586--34599. Curran Associates, Inc.

\bibitem[{Lewis et~al.(2020)Lewis, Perez, Piktus, Petroni, Karpukhin, Goyal, K{\"u}ttler, Lewis, Yih, Rockt{\"a}schel et~al.}]{lewis2020retrieval}
Patrick Lewis, Ethan Perez, Aleksandra Piktus, Fabio Petroni, Vladimir Karpukhin, Naman Goyal, Heinrich K{\"u}ttler, Mike Lewis, Wen-tau Yih, Tim Rockt{\"a}schel, et~al. 2020.
\newblock Retrieval-augmented generation for knowledge-intensive nlp tasks.
\newblock \emph{Advances in Neural Information Processing Systems}, 33:9459--9474.

\bibitem[{Li et~al.(2022)Li, Rawat, Zaheer, Wang, Lukasik, Veit, Yu, and Kumar}]{li2022large}
Daliang Li, Ankit~Singh Rawat, Manzil Zaheer, Xin Wang, Michal Lukasik, Andreas Veit, Felix Yu, and Sanjiv Kumar. 2022.
\newblock \href {http://arxiv.org/abs/2211.05110} {Large language models with controllable working memory}.

\bibitem[{Li et~al.(2023{\natexlab{a}})Li, Tang, Zhao, Wang, Nie, and Wen}]{li2023web}
Junyi Li, Tianyi Tang, Wayne~Xin Zhao, Jingyuan Wang, Jian-Yun Nie, and Ji-Rong Wen. 2023{\natexlab{a}}.
\newblock \href {http://arxiv.org/abs/2305.10998} {The web can be your oyster for improving large language models}.

\bibitem[{Li et~al.(2023{\natexlab{b}})Li, Li, Song, Yang, Ma, and Yu}]{li2023pmet}
Xiaopeng Li, Shasha Li, Shezheng Song, Jing Yang, Jun Ma, and Jie Yu. 2023{\natexlab{b}}.
\newblock \href {http://arxiv.org/abs/2308.08742} {Pmet: Precise model editing in a transformer}.

\bibitem[{Liang et~al.(2023)Liang, Wu, Song, Wu, Xia, Liu, Ou, Lu, Ji, Mao, Wang, Shou, Gong, and Duan}]{liang2023taskmatrixai}
Yaobo Liang, Chenfei Wu, Ting Song, Wenshan Wu, Yan Xia, Yu~Liu, Yang Ou, Shuai Lu, Lei Ji, Shaoguang Mao, Yun Wang, Linjun Shou, Ming Gong, and Nan Duan. 2023.
\newblock \href {http://arxiv.org/abs/2303.16434} {Taskmatrix.ai: Completing tasks by connecting foundation models with millions of apis}.

\bibitem[{Lin et~al.(2022)Lin, Wang, Lin, Jia, Xiao, Ren, and Yih}]{lin-etal-2022-continual}
Bill~Yuchen Lin, Sida Wang, Xi~Lin, Robin Jia, Lin Xiao, Xiang Ren, and Scott Yih. 2022.
\newblock \href {https://doi.org/10.18653/v1/2022.acl-long.223} {On continual model refinement in out-of-distribution data streams}.
\newblock In \emph{Proceedings of the 60th Annual Meeting of the Association for Computational Linguistics (Volume 1: Long Papers)}, pages 3128--3139, Dublin, Ireland. Association for Computational Linguistics.

\bibitem[{Liu et~al.(2023{\natexlab{a}})Liu, Zhang, and Liang}]{liu2023evaluating}
Nelson~F. Liu, Tianyi Zhang, and Percy Liang. 2023{\natexlab{a}}.
\newblock \href {http://arxiv.org/abs/2304.09848} {Evaluating verifiability in generative search engines}.

\bibitem[{Liu et~al.(2023{\natexlab{b}})Liu, Yuan, Fu, Jiang, Hayashi, and Neubig}]{10.1145/3560815}
Pengfei Liu, Weizhe Yuan, Jinlan Fu, Zhengbao Jiang, Hiroaki Hayashi, and Graham Neubig. 2023{\natexlab{b}}.
\newblock \href {https://doi.org/10.1145/3560815} {Pre-train, prompt, and predict: A systematic survey of prompting methods in natural language processing}.
\newblock \emph{ACM Comput. Surv.}, 55(9).

\bibitem[{Liu and Low(2023)}]{liu2023goat}
Tiedong Liu and Bryan Kian~Hsiang Low. 2023.
\newblock \href {http://arxiv.org/abs/2305.14201} {Goat: Fine-tuned llama outperforms gpt-4 on arithmetic tasks}.

\bibitem[{Liška et~al.(2022)Liška, Kočiský, Gribovskaya, Terzi, Sezener, Agrawal, de~Masson~d'Autume, Scholtes, Zaheer, Young, Gilsenan-McMahon, Austin, Blunsom, and Lazaridou}]{liška2022streamingqa}
Adam Liška, Tomáš Kočiský, Elena Gribovskaya, Tayfun Terzi, Eren Sezener, Devang Agrawal, Cyprien de~Masson~d'Autume, Tim Scholtes, Manzil Zaheer, Susannah Young, Ellen Gilsenan-McMahon, Sophia Austin, Phil Blunsom, and Angeliki Lazaridou. 2022.
\newblock \href {http://arxiv.org/abs/2205.11388} {Streamingqa: A benchmark for adaptation to new knowledge over time in question answering models}.

\bibitem[{Lu et~al.(2023)Lu, Peng, Cheng, Galley, Chang, Wu, Zhu, and Gao}]{lu2023chameleon}
Pan Lu, Baolin Peng, Hao Cheng, Michel Galley, Kai-Wei Chang, Ying~Nian Wu, Song-Chun Zhu, and Jianfeng Gao. 2023.
\newblock \href {http://arxiv.org/abs/2304.09842} {Chameleon: Plug-and-play compositional reasoning with large language models}.

\bibitem[{Luo et~al.(2023)Luo, Chuang, Gong, Zhang, Kim, Wu, Fox, Meng, and Glass}]{luo2023sail}
Hongyin Luo, Yung-Sung Chuang, Yuan Gong, Tianhua Zhang, Yoon Kim, Xixin Wu, Danny Fox, Helen Meng, and James Glass. 2023.
\newblock \href {http://arxiv.org/abs/2305.15225} {Sail: Search-augmented instruction learning}.

\bibitem[{Luu et~al.(2022)Luu, Khashabi, Gururangan, Mandyam, and Smith}]{luu-etal-2022-time}
Kelvin Luu, Daniel Khashabi, Suchin Gururangan, Karishma Mandyam, and Noah~A. Smith. 2022.
\newblock \href {https://doi.org/10.18653/v1/2022.naacl-main.435} {Time waits for no one! analysis and challenges of temporal misalignment}.
\newblock In \emph{Proceedings of the 2022 Conference of the North American Chapter of the Association for Computational Linguistics: Human Language Technologies}, pages 5944--5958, Seattle, United States. Association for Computational Linguistics.

\bibitem[{Ma et~al.(2023)Ma, Gong, He, Zhao, and Duan}]{ma2023query}
Xinbei Ma, Yeyun Gong, Pengcheng He, Hai Zhao, and Nan Duan. 2023.
\newblock \href {http://arxiv.org/abs/2305.14283} {Query rewriting for retrieval-augmented large language models}.

\bibitem[{Madaan et~al.(2022)Madaan, Tandon, Clark, and Yang}]{madaan-etal-2022-memory}
Aman Madaan, Niket Tandon, Peter Clark, and Yiming Yang. 2022.
\newblock \href {https://aclanthology.org/2022.emnlp-main.183} {Memory-assisted prompt editing to improve {GPT}-3 after deployment}.
\newblock In \emph{Proceedings of the 2022 Conference on Empirical Methods in Natural Language Processing}, pages 2833--2861, Abu Dhabi, United Arab Emirates. Association for Computational Linguistics.

\bibitem[{Mallen et~al.(2023)Mallen, Asai, Zhong, Das, Khashabi, and Hajishirzi}]{mallen2023trust}
Alex Mallen, Akari Asai, Victor Zhong, Rajarshi Das, Daniel Khashabi, and Hannaneh Hajishirzi. 2023.
\newblock \href {http://arxiv.org/abs/2212.10511} {When not to trust language models: Investigating effectiveness of parametric and non-parametric memories}.

\bibitem[{Meng et~al.(2022{\natexlab{a}})Meng, Bau, Andonian, and Belinkov}]{NEURIPS2022_6f1d43d5}
Kevin Meng, David Bau, Alex Andonian, and Yonatan Belinkov. 2022{\natexlab{a}}.
\newblock \href {https://proceedings.neurips.cc/paper_files/paper/2022/file/6f1d43d5a82a37e89b0665b33bf3a182-Paper-Conference.pdf} {Locating and editing factual associations in gpt}.
\newblock In \emph{Advances in Neural Information Processing Systems}, volume~35, pages 17359--17372. Curran Associates, Inc.

\bibitem[{Meng et~al.(2023)Meng, Sharma, Andonian, Belinkov, and Bau}]{meng2023massediting}
Kevin Meng, Arnab~Sen Sharma, Alex~J Andonian, Yonatan Belinkov, and David Bau. 2023.
\newblock \href {https://openreview.net/forum?id=MkbcAHIYgyS} {Mass-editing memory in a transformer}.
\newblock In \emph{The Eleventh International Conference on Learning Representations}.

\bibitem[{Meng et~al.(2022{\natexlab{b}})Meng, Zong, Li, Sun, Zhang, Wu, and Li}]{meng2022gnnlm}
Yuxian Meng, Shi Zong, Xiaoya Li, Xiaofei Sun, Tianwei Zhang, Fei Wu, and Jiwei Li. 2022{\natexlab{b}}.
\newblock \href {https://openreview.net/forum?id=BS49l-B5Bql} {{GNN}-{LM}: Language modeling based on global contexts via {GNN}}.
\newblock In \emph{International Conference on Learning Representations}.

\bibitem[{Menick et~al.(2022)Menick, Trebacz, Mikulik, Aslanides, Song, Chadwick, Glaese, Young, Campbell-Gillingham, Irving, and McAleese}]{menick2022teaching}
Jacob Menick, Maja Trebacz, Vladimir Mikulik, John Aslanides, Francis Song, Martin Chadwick, Mia Glaese, Susannah Young, Lucy Campbell-Gillingham, Geoffrey Irving, and Nat McAleese. 2022.
\newblock \href {http://arxiv.org/abs/2203.11147} {Teaching language models to support answers with verified quotes}.

\bibitem[{Mialon et~al.(2023)Mialon, Dessì, Lomeli, Nalmpantis, Pasunuru, Raileanu, Rozière, Schick, Dwivedi-Yu, Celikyilmaz, Grave, LeCun, and Scialom}]{mialon2023augmented}
Grégoire Mialon, Roberto Dessì, Maria Lomeli, Christoforos Nalmpantis, Ram Pasunuru, Roberta Raileanu, Baptiste Rozière, Timo Schick, Jane Dwivedi-Yu, Asli Celikyilmaz, Edouard Grave, Yann LeCun, and Thomas Scialom. 2023.
\newblock \href {http://arxiv.org/abs/2302.07842} {Augmented language models: a survey}.

\bibitem[{Mitchell et~al.(2022{\natexlab{a}})Mitchell, Lin, Bosselut, Finn, and Manning}]{mitchell2022fast}
Eric Mitchell, Charles Lin, Antoine Bosselut, Chelsea Finn, and Christopher~D Manning. 2022{\natexlab{a}}.
\newblock \href {https://openreview.net/forum?id=0DcZxeWfOPt} {Fast model editing at scale}.
\newblock In \emph{International Conference on Learning Representations}.

\bibitem[{Mitchell et~al.(2022{\natexlab{b}})Mitchell, Lin, Bosselut, Manning, and Finn}]{pmlr-v162-mitchell22a}
Eric Mitchell, Charles Lin, Antoine Bosselut, Christopher~D Manning, and Chelsea Finn. 2022{\natexlab{b}}.
\newblock \href {https://proceedings.mlr.press/v162/mitchell22a.html} {Memory-based model editing at scale}.
\newblock In \emph{Proceedings of the 39th International Conference on Machine Learning}, volume 162 of \emph{Proceedings of Machine Learning Research}, pages 15817--15831. PMLR.

\bibitem[{Nakano et~al.(2022)Nakano, Hilton, Balaji, Wu, Ouyang, Kim, Hesse, Jain, Kosaraju, Saunders, Jiang, Cobbe, Eloundou, Krueger, Button, Knight, Chess, and Schulman}]{nakano2022webgpt}
Reiichiro Nakano, Jacob Hilton, Suchir Balaji, Jeff Wu, Long Ouyang, Christina Kim, Christopher Hesse, Shantanu Jain, Vineet Kosaraju, William Saunders, Xu~Jiang, Karl Cobbe, Tyna Eloundou, Gretchen Krueger, Kevin Button, Matthew Knight, Benjamin Chess, and John Schulman. 2022.
\newblock \href {http://arxiv.org/abs/2112.09332} {Webgpt: Browser-assisted question-answering with human feedback}.

\bibitem[{Neeman et~al.(2022)Neeman, Aharoni, Honovich, Choshen, Szpektor, and Abend}]{neeman2022disentqa}
Ella Neeman, Roee Aharoni, Or~Honovich, Leshem Choshen, Idan Szpektor, and Omri Abend. 2022.
\newblock \href {http://arxiv.org/abs/2211.05655} {Disentqa: Disentangling parametric and contextual knowledge with counterfactual question answering}.

\bibitem[{Onoe et~al.(2023)Onoe, Zhang, Padmanabhan, Durrett, and Choi}]{onoe2023lms}
Yasumasa Onoe, Michael J.~Q. Zhang, Shankar Padmanabhan, Greg Durrett, and Eunsol Choi. 2023.
\newblock \href {http://arxiv.org/abs/2305.01651} {Can lms learn new entities from descriptions? challenges in propagating injected knowledge}.

\bibitem[{OpenAI(2022)}]{OpenAI_chatgpt2022}
OpenAI. 2022.
\newblock \href {https://openai.com/blog/chatgpt} {Introducing chatgpt}.

\bibitem[{OpenAI(2023{\natexlab{a}})}]{OpenAI_plugin2023}
OpenAI. 2023{\natexlab{a}}.
\newblock \href {https://openai.com/blog/chatgpt-plugins} {chatgpt plugins}.

\bibitem[{OpenAI(2023{\natexlab{b}})}]{openai2023gpt4}
OpenAI. 2023{\natexlab{b}}.
\newblock \href {http://arxiv.org/abs/2303.08774} {Gpt-4 technical report}.

\bibitem[{Ouyang et~al.(2022)Ouyang, Wu, Jiang, Almeida, Wainwright, Mishkin, Zhang, Agarwal, Slama, Ray, Schulman, Hilton, Kelton, Miller, Simens, Askell, Welinder, Christiano, Leike, and Lowe}]{NEURIPS2022_b1efde53}
Long Ouyang, Jeffrey Wu, Xu~Jiang, Diogo Almeida, Carroll Wainwright, Pamela Mishkin, Chong Zhang, Sandhini Agarwal, Katarina Slama, Alex Ray, John Schulman, Jacob Hilton, Fraser Kelton, Luke Miller, Maddie Simens, Amanda Askell, Peter Welinder, Paul~F Christiano, Jan Leike, and Ryan Lowe. 2022.
\newblock \href {https://proceedings.neurips.cc/paper_files/paper/2022/file/b1efde53be364a73914f58805a001731-Paper-Conference.pdf} {Training language models to follow instructions with human feedback}.
\newblock In \emph{Advances in Neural Information Processing Systems}, volume~35, pages 27730--27744. Curran Associates, Inc.

\bibitem[{Padmanabhan et~al.(2023)Padmanabhan, Onoe, Zhang, Durrett, and Choi}]{padmanabhan2023propagating}
Shankar Padmanabhan, Yasumasa Onoe, Michael J.~Q. Zhang, Greg Durrett, and Eunsol Choi. 2023.
\newblock \href {http://arxiv.org/abs/2306.09306} {Propagating knowledge updates to lms through distillation}.

\bibitem[{Pan et~al.(2023)Pan, Luo, Wang, Chen, Wang, and Wu}]{pan2023unifying}
Shirui Pan, Linhao Luo, Yufei Wang, Chen Chen, Jiapu Wang, and Xindong Wu. 2023.
\newblock \href {http://arxiv.org/abs/2306.08302} {Unifying large language models and knowledge graphs: A roadmap}.

\bibitem[{Paranjape et~al.(2023)Paranjape, Lundberg, Singh, Hajishirzi, Zettlemoyer, and Ribeiro}]{paranjape2023art}
Bhargavi Paranjape, Scott Lundberg, Sameer Singh, Hannaneh Hajishirzi, Luke Zettlemoyer, and Marco~Tulio Ribeiro. 2023.
\newblock \href {http://arxiv.org/abs/2303.09014} {Art: Automatic multi-step reasoning and tool-use for large language models}.

\bibitem[{Patterson et~al.(2021)Patterson, Gonzalez, Le, Liang, Munguia, Rothchild, So, Texier, and Dean}]{patterson2021carbon}
David Patterson, Joseph Gonzalez, Quoc Le, Chen Liang, Lluis-Miquel Munguia, Daniel Rothchild, David So, Maud Texier, and Jeff Dean. 2021.
\newblock \href {http://arxiv.org/abs/2104.10350} {Carbon emissions and large neural network training}.

\bibitem[{Peng et~al.(2023{\natexlab{a}})Peng, Galley, He, Cheng, Xie, Hu, Huang, Liden, Yu, Chen, and Gao}]{peng2023check}
Baolin Peng, Michel Galley, Pengcheng He, Hao Cheng, Yujia Xie, Yu~Hu, Qiuyuan Huang, Lars Liden, Zhou Yu, Weizhu Chen, and Jianfeng Gao. 2023{\natexlab{a}}.
\newblock \href {http://arxiv.org/abs/2302.12813} {Check your facts and try again: Improving large language models with external knowledge and automated feedback}.

\bibitem[{Peng et~al.(2023{\natexlab{b}})Peng, Ge, Chen, Wei, and Wang}]{peng2023semiparametric}
Guangyue Peng, Tao Ge, Si-Qing Chen, Furu Wei, and Houfeng Wang. 2023{\natexlab{b}}.
\newblock \href {http://arxiv.org/abs/2303.01421} {Semiparametric language models are scalable continual learners}.

\bibitem[{Petroni et~al.(2019)Petroni, Rockt{\"a}schel, Riedel, Lewis, Bakhtin, Wu, and Miller}]{petroni-etal-2019-language}
Fabio Petroni, Tim Rockt{\"a}schel, Sebastian Riedel, Patrick Lewis, Anton Bakhtin, Yuxiang Wu, and Alexander Miller. 2019.
\newblock \href {https://doi.org/10.18653/v1/D19-1250} {Language models as knowledge bases?}
\newblock In \emph{Proceedings of the 2019 Conference on Empirical Methods in Natural Language Processing and the 9th International Joint Conference on Natural Language Processing (EMNLP-IJCNLP)}, pages 2463--2473, Hong Kong, China. Association for Computational Linguistics.

\bibitem[{Press et~al.(2023)Press, Zhang, Min, Schmidt, Smith, and Lewis}]{press2023measuring}
Ofir Press, Muru Zhang, Sewon Min, Ludwig Schmidt, Noah~A. Smith, and Mike Lewis. 2023.
\newblock \href {http://arxiv.org/abs/2210.03350} {Measuring and narrowing the compositionality gap in language models}.

\bibitem[{Qin et~al.(2023{\natexlab{a}})Qin, Cai, Jin, Yan, Liang, Zhu, Lin, Han, Ding, Wang, Xie, Qi, Liu, Sun, and Zhou}]{qin2023webcpm}
Yujia Qin, Zihan Cai, Dian Jin, Lan Yan, Shihao Liang, Kunlun Zhu, Yankai Lin, Xu~Han, Ning Ding, Huadong Wang, Ruobing Xie, Fanchao Qi, Zhiyuan Liu, Maosong Sun, and Jie Zhou. 2023{\natexlab{a}}.
\newblock \href {http://arxiv.org/abs/2305.06849} {Webcpm: Interactive web search for chinese long-form question answering}.

\bibitem[{Qin et~al.(2023{\natexlab{b}})Qin, Hu, Lin, Chen, Ding, Cui, Zeng, Huang, Xiao, Han, Fung, Su, Wang, Qian, Tian, Zhu, Liang, Shen, Xu, Zhang, Ye, Li, Tang, Yi, Zhu, Dai, Yan, Cong, Lu, Zhao, Huang, Yan, Han, Sun, Li, Phang, Yang, Wu, Ji, Liu, and Sun}]{qin2023tool}
Yujia Qin, Shengding Hu, Yankai Lin, Weize Chen, Ning Ding, Ganqu Cui, Zheni Zeng, Yufei Huang, Chaojun Xiao, Chi Han, Yi~Ren Fung, Yusheng Su, Huadong Wang, Cheng Qian, Runchu Tian, Kunlun Zhu, Shihao Liang, Xingyu Shen, Bokai Xu, Zhen Zhang, Yining Ye, Bowen Li, Ziwei Tang, Jing Yi, Yuzhang Zhu, Zhenning Dai, Lan Yan, Xin Cong, Yaxi Lu, Weilin Zhao, Yuxiang Huang, Junxi Yan, Xu~Han, Xian Sun, Dahai Li, Jason Phang, Cheng Yang, Tongshuang Wu, Heng Ji, Zhiyuan Liu, and Maosong Sun. 2023{\natexlab{b}}.
\newblock \href {http://arxiv.org/abs/2304.08354} {Tool learning with foundation models}.

\bibitem[{Qin et~al.(2022)Qin, Zhang, Lin, Liu, Li, Sun, and Zhou}]{qin-etal-2022-elle}
Yujia Qin, Jiajie Zhang, Yankai Lin, Zhiyuan Liu, Peng Li, Maosong Sun, and Jie Zhou. 2022.
\newblock \href {https://doi.org/10.18653/v1/2022.findings-acl.220} {{ELLE}: Efficient lifelong pre-training for emerging data}.
\newblock In \emph{Findings of the Association for Computational Linguistics: ACL 2022}, pages 2789--2810, Dublin, Ireland. Association for Computational Linguistics.

\bibitem[{Radford et~al.(2018)Radford, Narasimhan, Salimans, Sutskever et~al.}]{radford2018improving}
Alec Radford, Karthik Narasimhan, Tim Salimans, Ilya Sutskever, et~al. 2018.
\newblock Improving language understanding by generative pre-training.

\bibitem[{Radford et~al.(2019)Radford, Wu, Child, Luan, Amodei, and Sutskever}]{radford2019language}
Alec Radford, Jeff Wu, Rewon Child, David Luan, Dario Amodei, and Ilya Sutskever. 2019.
\newblock Language models are unsupervised multitask learners.

\bibitem[{Raffel et~al.(2020)Raffel, Shazeer, Roberts, Lee, Narang, Matena, Zhou, Li, and Liu}]{raffel2020exploring}
Colin Raffel, Noam Shazeer, Adam Roberts, Katherine Lee, Sharan Narang, Michael Matena, Yanqi Zhou, Wei Li, and Peter~J Liu. 2020.
\newblock Exploring the limits of transfer learning with a unified text-to-text transformer.
\newblock \emph{Journal of Machine Learning Research}, 21:1--67.

\bibitem[{Ram et~al.(2023)Ram, Levine, Dalmedigos, Muhlgay, Shashua, Leyton-Brown, and Shoham}]{ram2023incontext}
Ori Ram, Yoav Levine, Itay Dalmedigos, Dor Muhlgay, Amnon Shashua, Kevin Leyton-Brown, and Yoav Shoham. 2023.
\newblock \href {http://arxiv.org/abs/2302.00083} {In-context retrieval-augmented language models}.

\bibitem[{Roberts et~al.(2020)Roberts, Raffel, and Shazeer}]{roberts-etal-2020-much}
Adam Roberts, Colin Raffel, and Noam Shazeer. 2020.
\newblock \href {https://doi.org/10.18653/v1/2020.emnlp-main.437} {How much knowledge can you pack into the parameters of a language model?}
\newblock In \emph{Proceedings of the 2020 Conference on Empirical Methods in Natural Language Processing (EMNLP)}, pages 5418--5426, Online. Association for Computational Linguistics.

\bibitem[{R{\"o}ttger and Pierrehumbert(2021)}]{rottger-pierrehumbert-2021-temporal-adaptation}
Paul R{\"o}ttger and Janet Pierrehumbert. 2021.
\newblock \href {https://doi.org/10.18653/v1/2021.findings-emnlp.206} {Temporal adaptation of {BERT} and performance on downstream document classification: Insights from social media}.
\newblock In \emph{Findings of the Association for Computational Linguistics: EMNLP 2021}, pages 2400--2412, Punta Cana, Dominican Republic. Association for Computational Linguistics.

\bibitem[{Sanh et~al.(2022)Sanh, Webson, Raffel, Bach, Sutawika, Alyafeai, Chaffin, Stiegler, Raja, Dey, Bari, Xu, Thakker, Sharma, Szczechla, Kim, Chhablani, Nayak, Datta, Chang, Jiang, Wang, Manica, Shen, Yong, Pandey, Bawden, Wang, Neeraj, Rozen, Sharma, Santilli, Fevry, Fries, Teehan, Scao, Biderman, Gao, Wolf, and Rush}]{sanh2022multitask}
Victor Sanh, Albert Webson, Colin Raffel, Stephen Bach, Lintang Sutawika, Zaid Alyafeai, Antoine Chaffin, Arnaud Stiegler, Arun Raja, Manan Dey, M~Saiful Bari, Canwen Xu, Urmish Thakker, Shanya~Sharma Sharma, Eliza Szczechla, Taewoon Kim, Gunjan Chhablani, Nihal Nayak, Debajyoti Datta, Jonathan Chang, Mike Tian-Jian Jiang, Han Wang, Matteo Manica, Sheng Shen, Zheng~Xin Yong, Harshit Pandey, Rachel Bawden, Thomas Wang, Trishala Neeraj, Jos Rozen, Abheesht Sharma, Andrea Santilli, Thibault Fevry, Jason~Alan Fries, Ryan Teehan, Teven~Le Scao, Stella Biderman, Leo Gao, Thomas Wolf, and Alexander~M Rush. 2022.
\newblock \href {https://openreview.net/forum?id=9Vrb9D0WI4} {Multitask prompted training enables zero-shot task generalization}.
\newblock In \emph{International Conference on Learning Representations}.

\bibitem[{Scialom et~al.(2022)Scialom, Chakrabarty, and Muresan}]{scialom-etal-2022-fine}
Thomas Scialom, Tuhin Chakrabarty, and Smaranda Muresan. 2022.
\newblock \href {https://aclanthology.org/2022.emnlp-main.410} {Fine-tuned language models are continual learners}.
\newblock In \emph{Proceedings of the 2022 Conference on Empirical Methods in Natural Language Processing}, pages 6107--6122, Abu Dhabi, United Arab Emirates. Association for Computational Linguistics.

\bibitem[{Semnani et~al.(2023)Semnani, Yao, Zhang, and Lam}]{semnani2023wikichat}
Sina~J. Semnani, Violet~Z. Yao, Heidi~C. Zhang, and Monica~S. Lam. 2023.
\newblock \href {http://arxiv.org/abs/2305.14292} {Wikichat: A few-shot llm-based chatbot grounded with wikipedia}.

\bibitem[{Shao et~al.(2023)Shao, Gong, Shen, Huang, Duan, and Chen}]{shao2023enhancing}
Zhihong Shao, Yeyun Gong, Yelong Shen, Minlie Huang, Nan Duan, and Weizhu Chen. 2023.
\newblock \href {http://arxiv.org/abs/2305.15294} {Enhancing retrieval-augmented large language models with iterative retrieval-generation synergy}.

\bibitem[{Shen et~al.(2023)Shen, Zhang, Cao, Tan, Chen, and Gan}]{shen2023moduleformer}
Yikang Shen, Zheyu Zhang, Tianyou Cao, Shawn Tan, Zhenfang Chen, and Chuang Gan. 2023.
\newblock \href {http://arxiv.org/abs/2306.04640} {Moduleformer: Modularity emerges from mixture-of-experts}.

\bibitem[{Shi et~al.(2023{\natexlab{a}})Shi, Chen, Misra, Scales, Dohan, Chi, Schärli, and Zhou}]{shi2023large}
Freda Shi, Xinyun Chen, Kanishka Misra, Nathan Scales, David Dohan, Ed~Chi, Nathanael Schärli, and Denny Zhou. 2023{\natexlab{a}}.
\newblock \href {http://arxiv.org/abs/2302.00093} {Large language models can be easily distracted by irrelevant context}.

\bibitem[{Shi et~al.(2022)Shi, Michael, Gururangan, and Zettlemoyer}]{shi-etal-2022-nearest}
Weijia Shi, Julian Michael, Suchin Gururangan, and Luke Zettlemoyer. 2022.
\newblock \href {https://aclanthology.org/2022.emnlp-main.214} {Nearest neighbor zero-shot inference}.
\newblock In \emph{Proceedings of the 2022 Conference on Empirical Methods in Natural Language Processing}, pages 3254--3265, Abu Dhabi, United Arab Emirates. Association for Computational Linguistics.

\bibitem[{Shi et~al.(2023{\natexlab{b}})Shi, Min, Yasunaga, Seo, James, Lewis, Zettlemoyer, and tau Yih}]{shi2023replug}
Weijia Shi, Sewon Min, Michihiro Yasunaga, Minjoon Seo, Rich James, Mike Lewis, Luke Zettlemoyer, and Wen tau Yih. 2023{\natexlab{b}}.
\newblock \href {http://arxiv.org/abs/2301.12652} {Replug: Retrieval-augmented black-box language models}.

\bibitem[{Shuster et~al.(2022)Shuster, Xu, Komeili, Ju, Smith, Roller, Ung, Chen, Arora, Lane, Behrooz, Ngan, Poff, Goyal, Szlam, Boureau, Kambadur, and Weston}]{shuster2022blenderbot}
Kurt Shuster, Jing Xu, Mojtaba Komeili, Da~Ju, Eric~Michael Smith, Stephen Roller, Megan Ung, Moya Chen, Kushal Arora, Joshua Lane, Morteza Behrooz, William Ngan, Spencer Poff, Naman Goyal, Arthur Szlam, Y-Lan Boureau, Melanie Kambadur, and Jason Weston. 2022.
\newblock \href {http://arxiv.org/abs/2208.03188} {Blenderbot 3: a deployed conversational agent that continually learns to responsibly engage}.

\bibitem[{Si et~al.(2023)Si, Gan, Yang, Wang, Wang, Boyd-Graber, and Wang}]{si2023prompting}
Chenglei Si, Zhe Gan, Zhengyuan Yang, Shuohang Wang, Jianfeng Wang, Jordan~Lee Boyd-Graber, and Lijuan Wang. 2023.
\newblock \href {https://openreview.net/forum?id=98p5x51L5af} {Prompting {GPT}-3 to be reliable}.
\newblock In \emph{The Eleventh International Conference on Learning Representations}.

\bibitem[{Singhal et~al.(2022)Singhal, Azizi, Tu, Mahdavi, Wei, Chung, Scales, Tanwani, Cole-Lewis, Pfohl, Payne, Seneviratne, Gamble, Kelly, Scharli, Chowdhery, Mansfield, y~Arcas, Webster, Corrado, Matias, Chou, Gottweis, Tomasev, Liu, Rajkomar, Barral, Semturs, Karthikesalingam, and Natarajan}]{singhal2022large}
Karan Singhal, Shekoofeh Azizi, Tao Tu, S.~Sara Mahdavi, Jason Wei, Hyung~Won Chung, Nathan Scales, Ajay Tanwani, Heather Cole-Lewis, Stephen Pfohl, Perry Payne, Martin Seneviratne, Paul Gamble, Chris Kelly, Nathaneal Scharli, Aakanksha Chowdhery, Philip Mansfield, Blaise~Aguera y~Arcas, Dale Webster, Greg~S. Corrado, Yossi Matias, Katherine Chou, Juraj Gottweis, Nenad Tomasev, Yun Liu, Alvin Rajkomar, Joelle Barral, Christopher Semturs, Alan Karthikesalingam, and Vivek Natarajan. 2022.
\newblock \href {http://arxiv.org/abs/2212.13138} {Large language models encode clinical knowledge}.

\bibitem[{Sinitsin et~al.(2020)Sinitsin, Plokhotnyuk, Pyrkin, Popov, and Babenko}]{Sinitsin2020Editable}
Anton Sinitsin, Vsevolod Plokhotnyuk, Dmitry Pyrkin, Sergei Popov, and Artem Babenko. 2020.
\newblock \href {https://openreview.net/forum?id=HJedXaEtvS} {Editable neural networks}.
\newblock In \emph{International Conference on Learning Representations}.

\bibitem[{Tandon et~al.(2022)Tandon, Madaan, Clark, and Yang}]{tandon-etal-2022-learning}
Niket Tandon, Aman Madaan, Peter Clark, and Yiming Yang. 2022.
\newblock \href {https://doi.org/10.18653/v1/2022.findings-naacl.26} {Learning to repair: Repairing model output errors after deployment using a dynamic memory of feedback}.
\newblock In \emph{Findings of the Association for Computational Linguistics: NAACL 2022}, pages 339--352, Seattle, United States. Association for Computational Linguistics.

\bibitem[{Tirumala et~al.(2022)Tirumala, Markosyan, Zettlemoyer, and Aghajanyan}]{NEURIPS2022_fa0509f4}
Kushal Tirumala, Aram Markosyan, Luke Zettlemoyer, and Armen Aghajanyan. 2022.
\newblock \href {https://proceedings.neurips.cc/paper_files/paper/2022/file/fa0509f4dab6807e2cb465715bf2d249-Paper-Conference.pdf} {Memorization without overfitting: Analyzing the training dynamics of large language models}.
\newblock In \emph{Advances in Neural Information Processing Systems}, volume~35, pages 38274--38290. Curran Associates, Inc.

\bibitem[{Touvron et~al.(2023)Touvron, Lavril, Izacard, Martinet, Lachaux, Lacroix, Rozière, Goyal, Hambro, Azhar, Rodriguez, Joulin, Grave, and Lample}]{touvron2023llama}
Hugo Touvron, Thibaut Lavril, Gautier Izacard, Xavier Martinet, Marie-Anne Lachaux, Timothée Lacroix, Baptiste Rozière, Naman Goyal, Eric Hambro, Faisal Azhar, Aurelien Rodriguez, Armand Joulin, Edouard Grave, and Guillaume Lample. 2023.
\newblock \href {http://arxiv.org/abs/2302.13971} {Llama: Open and efficient foundation language models}.

\bibitem[{Trivedi et~al.(2022)Trivedi, Balasubramanian, Khot, and Sabharwal}]{trivedi2022interleaving}
Harsh Trivedi, Niranjan Balasubramanian, Tushar Khot, and Ashish Sabharwal. 2022.
\newblock \href {http://arxiv.org/abs/2212.10509} {Interleaving retrieval with chain-of-thought reasoning for knowledge-intensive multi-step questions}.

\bibitem[{Vaswani et~al.(2017)Vaswani, Shazeer, Parmar, Uszkoreit, Jones, Gomez, Kaiser, and Polosukhin}]{NIPS2017_3f5ee243}
Ashish Vaswani, Noam Shazeer, Niki Parmar, Jakob Uszkoreit, Llion Jones, Aidan~N Gomez, \L~ukasz Kaiser, and Illia Polosukhin. 2017.
\newblock \href {https://proceedings.neurips.cc/paper_files/paper/2017/file/3f5ee243547dee91fbd053c1c4a845aa-Paper.pdf} {Attention is all you need}.
\newblock In \emph{Advances in Neural Information Processing Systems}, volume~30. Curran Associates, Inc.

\bibitem[{Wang et~al.(2023{\natexlab{a}})Wang, Liang, Sun, Cao, and Xu}]{wang2023crosslingual}
Jiaan Wang, Yunlong Liang, Zengkui Sun, Yuxuan Cao, and Jiarong Xu. 2023{\natexlab{a}}.
\newblock \href {http://arxiv.org/abs/2309.08952} {Cross-lingual knowledge editing in large language models}.

\bibitem[{Wang et~al.(2021)Wang, Tang, Duan, Wei, Huang, Ji, Cao, Jiang, and Zhou}]{wang-etal-2021-k}
Ruize Wang, Duyu Tang, Nan Duan, Zhongyu Wei, Xuanjing Huang, Jianshu Ji, Guihong Cao, Daxin Jiang, and Ming Zhou. 2021.
\newblock \href {https://doi.org/10.18653/v1/2021.findings-acl.121} {{K-Adapter}: {I}nfusing {K}nowledge into {P}re-{T}rained {M}odels with {A}dapters}.
\newblock In \emph{Findings of the Association for Computational Linguistics: ACL-IJCNLP 2021}, pages 1405--1418, Online. Association for Computational Linguistics.

\bibitem[{Wang et~al.(2023{\natexlab{b}})Wang, Zhang, Yang, Shi, Zhou, Hao, Xiong, Li, Sim, Chen, Zhu, Yang, Nik, Liu, Lin, Wang, Liu, Chen, Xu, Liu, Guo, and Fu}]{wang2023interactive}
Zekun Wang, Ge~Zhang, Kexin Yang, Ning Shi, Wangchunshu Zhou, Shaochun Hao, Guangzheng Xiong, Yizhi Li, Mong~Yuan Sim, Xiuying Chen, Qingqing Zhu, Zhenzhu Yang, Adam Nik, Qi~Liu, Chenghua Lin, Shi Wang, Ruibo Liu, Wenhu Chen, Ke~Xu, Dayiheng Liu, Yike Guo, and Jie Fu. 2023{\natexlab{b}}.
\newblock \href {http://arxiv.org/abs/2305.13246} {Interactive natural language processing}.

\bibitem[{Wang et~al.(2023{\natexlab{c}})Wang, Yang, Shen, and Huang}]{wang2023comprehensive}
Zhenyi Wang, Enneng Yang, Li~Shen, and Heng Huang. 2023{\natexlab{c}}.
\newblock \href {http://arxiv.org/abs/2307.09218} {A comprehensive survey of forgetting in deep learning beyond continual learning}.

\bibitem[{Wei et~al.(2022)Wei, Wang, Schuurmans, Bosma, brian ichter, Xia, Chi, Le, and Zhou}]{wei2022chain}
Jason Wei, Xuezhi Wang, Dale Schuurmans, Maarten Bosma, brian ichter, Fei Xia, Ed~H. Chi, Quoc~V Le, and Denny Zhou. 2022.
\newblock \href {https://openreview.net/forum?id=_VjQlMeSB_J} {Chain of thought prompting elicits reasoning in large language models}.
\newblock In \emph{Advances in Neural Information Processing Systems}.

\bibitem[{Wei et~al.(2021)Wei, Wang, Zhang, Bhatia, and Arnold}]{wei2021knowledge}
Xiaokai Wei, Shen Wang, Dejiao Zhang, Parminder Bhatia, and Andrew Arnold. 2021.
\newblock \href {http://arxiv.org/abs/2110.08455} {Knowledge enhanced pretrained language models: A compreshensive survey}.

\bibitem[{Wu et~al.(2023)Wu, Peng, Chen, Su, and Sun}]{wu2023evakellm}
Suhang Wu, Minlong Peng, Yue Chen, Jinsong Su, and Mingming Sun. 2023.
\newblock \href {http://arxiv.org/abs/2308.09954} {Eva-kellm: A new benchmark for evaluating knowledge editing of llms}.

\bibitem[{Wu et~al.(2022)Wu, Rabe, Hutchins, and Szegedy}]{wu2022memorizing}
Yuhuai Wu, Markus~Norman Rabe, DeLesley Hutchins, and Christian Szegedy. 2022.
\newblock \href {https://openreview.net/forum?id=TrjbxzRcnf-} {Memorizing transformers}.
\newblock In \emph{International Conference on Learning Representations}.

\bibitem[{Xie et~al.(2023)Xie, Zhang, Chen, Lou, and Su}]{xie2023adaptive}
Jian Xie, Kai Zhang, Jiangjie Chen, Renze Lou, and Yu~Su. 2023.
\newblock \href {http://arxiv.org/abs/2305.13300} {Adaptive chameleon or stubborn sloth: Unraveling the behavior of large language models in knowledge clashes}.

\bibitem[{Xu et~al.(2023{\natexlab{a}})Xu, Namazifar, Hazarika, Padmakumar, Liu, and Hakkani-Tür}]{xu2023kilm}
Yan Xu, Mahdi Namazifar, Devamanyu Hazarika, Aishwarya Padmakumar, Yang Liu, and Dilek Hakkani-Tür. 2023{\natexlab{a}}.
\newblock \href {http://arxiv.org/abs/2302.09170} {Kilm: Knowledge injection into encoder-decoder language models}.

\bibitem[{Xu et~al.(2023{\natexlab{b}})Xu, Hou, Che, and Zhang}]{xu2023language}
Yang Xu, Yutai Hou, Wanxiang Che, and Min Zhang. 2023{\natexlab{b}}.
\newblock \href {http://arxiv.org/abs/2205.12677} {Language anisotropic cross-lingual model editing}.

\bibitem[{Yang et~al.(2023)Yang, Li, Wang, Lin, Azarnasab, Ahmed, Liu, Liu, Zeng, and Wang}]{yang2023mmreact}
Zhengyuan Yang, Linjie Li, Jianfeng Wang, Kevin Lin, Ehsan Azarnasab, Faisal Ahmed, Zicheng Liu, Ce~Liu, Michael Zeng, and Lijuan Wang. 2023.
\newblock \href {http://arxiv.org/abs/2303.11381} {Mm-react: Prompting chatgpt for multimodal reasoning and action}.

\bibitem[{Yao et~al.(2023{\natexlab{a}})Yao, Zhao, Yu, Du, Shafran, Narasimhan, and Cao}]{yao2023react}
Shunyu Yao, Jeffrey Zhao, Dian Yu, Nan Du, Izhak Shafran, Karthik~R Narasimhan, and Yuan Cao. 2023{\natexlab{a}}.
\newblock \href {https://openreview.net/forum?id=WE_vluYUL-X} {React: Synergizing reasoning and acting in language models}.
\newblock In \emph{The Eleventh International Conference on Learning Representations}.

\bibitem[{Yao et~al.(2023{\natexlab{b}})Yao, Wang, Tian, Cheng, Li, Deng, Chen, and Zhang}]{yao2023editing}
Yunzhi Yao, Peng Wang, Bozhong Tian, Siyuan Cheng, Zhoubo Li, Shumin Deng, Huajun Chen, and Ningyu Zhang. 2023{\natexlab{b}}.
\newblock \href {http://arxiv.org/abs/2305.13172} {Editing large language models: Problems, methods, and opportunities}.

\bibitem[{Yin et~al.(2022)Yin, Dong, Cheng, Liu, Chang, Wei, and Gao}]{yin2022survey}
Da~Yin, Li~Dong, Hao Cheng, Xiaodong Liu, Kai-Wei Chang, Furu Wei, and Jianfeng Gao. 2022.
\newblock \href {http://arxiv.org/abs/2202.08772} {A survey of knowledge-intensive nlp with pre-trained language models}.

\bibitem[{Yin et~al.(2023)Yin, Sun, Guo, Wu, Qiu, and Huang}]{yin2023large}
Zhangyue Yin, Qiushi Sun, Qipeng Guo, Jiawen Wu, Xipeng Qiu, and Xuanjing Huang. 2023.
\newblock \href {http://arxiv.org/abs/2305.18153} {Do large language models know what they don't know?}

\bibitem[{Yu and Ji(2023)}]{yu2023self}
Pengfei Yu and Heng Ji. 2023.
\newblock \href {http://arxiv.org/abs/2305.18582} {Self information update for large language models through mitigating exposure bias}.

\bibitem[{Yu et~al.(2023{\natexlab{a}})Yu, Zhang, Liang, Jiang, and Sabharwal}]{yu2023improving}
Wenhao Yu, Zhihan Zhang, Zhenwen Liang, Meng Jiang, and Ashish Sabharwal. 2023{\natexlab{a}}.
\newblock \href {http://arxiv.org/abs/2305.14002} {Improving language models via plug-and-play retrieval feedback}.

\bibitem[{Yu et~al.(2022)Yu, Zhu, Li, Hu, Wang, Ji, and Jiang}]{10.1145/3512467}
Wenhao Yu, Chenguang Zhu, Zaitang Li, Zhiting Hu, Qingyun Wang, Heng Ji, and Meng Jiang. 2022.
\newblock \href {https://doi.org/10.1145/3512467} {A survey of knowledge-enhanced text generation}.
\newblock \emph{ACM Comput. Surv.}, 54(11s).

\bibitem[{Yu et~al.(2023{\natexlab{b}})Yu, Xiong, Yu, and Liu}]{yu2023augmentationadapted}
Zichun Yu, Chenyan Xiong, Shi Yu, and Zhiyuan Liu. 2023{\natexlab{b}}.
\newblock \href {http://arxiv.org/abs/2305.17331} {Augmentation-adapted retriever improves generalization of language models as generic plug-in}.

\bibitem[{Zhang and Choi(2023)}]{zhang2023mitigating}
Michael J.~Q. Zhang and Eunsol Choi. 2023.
\newblock \href {http://arxiv.org/abs/2305.14824} {Mitigating temporal misalignment by discarding outdated facts}.

\bibitem[{Zhang et~al.(2022)Zhang, Roller, Goyal, Artetxe, Chen, Chen, Dewan, Diab, Li, Lin, Mihaylov, Ott, Shleifer, Shuster, Simig, Koura, Sridhar, Wang, and Zettlemoyer}]{zhang2022opt}
Susan Zhang, Stephen Roller, Naman Goyal, Mikel Artetxe, Moya Chen, Shuohui Chen, Christopher Dewan, Mona Diab, Xian Li, Xi~Victoria Lin, Todor Mihaylov, Myle Ott, Sam Shleifer, Kurt Shuster, Daniel Simig, Punit~Singh Koura, Anjali Sridhar, Tianlu Wang, and Luke Zettlemoyer. 2022.
\newblock \href {http://arxiv.org/abs/2205.01068} {Opt: Open pre-trained transformer language models}.

\bibitem[{Zhang et~al.(2023)Zhang, Luo, Chuang, Fang, Gaitskell, Hartvigsen, Wu, Fox, Meng, and Glass}]{zhang2023interpretable}
Tianhua Zhang, Hongyin Luo, Yung-Sung Chuang, Wei Fang, Luc Gaitskell, Thomas Hartvigsen, Xixin Wu, Danny Fox, Helen Meng, and James Glass. 2023.
\newblock \href {http://arxiv.org/abs/2304.03728} {Interpretable unified language checking}.

\bibitem[{Zhao et~al.(2023)Zhao, Li, Joty, Qin, and Bing}]{zhao2023verifyandedit}
Ruochen Zhao, Xingxuan Li, Shafiq Joty, Chengwei Qin, and Lidong Bing. 2023.
\newblock \href {http://arxiv.org/abs/2305.03268} {Verify-and-edit: A knowledge-enhanced chain-of-thought framework}.

\bibitem[{Zhen et~al.(2022)Zhen, Shang, Liu, Li, Chen, and Zhang}]{zhen2022survey}
Chaoqi Zhen, Yanlei Shang, Xiangyu Liu, Yifei Li, Yong Chen, and Dell Zhang. 2022.
\newblock \href {http://arxiv.org/abs/2212.13428} {A survey on knowledge-enhanced pre-trained language models}.

\bibitem[{Zheng et~al.(2023)Zheng, Li, Dong, Fan, Wu, Xu, and Chang}]{zheng2023edit}
Ce~Zheng, Lei Li, Qingxiu Dong, Yuxuan Fan, Zhiyong Wu, Jingjing Xu, and Baobao Chang. 2023.
\newblock \href {http://arxiv.org/abs/2305.12740} {Can we edit factual knowledge by in-context learning?}

\bibitem[{Zhong et~al.(2023)Zhong, Wu, Manning, Potts, and Chen}]{zhong2023mquake}
Zexuan Zhong, Zhengxuan Wu, Christopher~D. Manning, Christopher Potts, and Danqi Chen. 2023.
\newblock \href {http://arxiv.org/abs/2305.14795} {Mquake: Assessing knowledge editing in language models via multi-hop questions}.

\bibitem[{Zhou et~al.(2023)Zhou, Zhang, Poon, and Chen}]{zhou2023contextfaithful}
Wenxuan Zhou, Sheng Zhang, Hoifung Poon, and Muhao Chen. 2023.
\newblock \href {http://arxiv.org/abs/2303.11315} {Context-faithful prompting for large language models}.

\bibitem[{Zhu et~al.(2020)Zhu, Rawat, Zaheer, Bhojanapalli, Li, Yu, and Kumar}]{zhu2020modifying}
Chen Zhu, Ankit~Singh Rawat, Manzil Zaheer, Srinadh Bhojanapalli, Daliang Li, Felix Yu, and Sanjiv Kumar. 2020.
\newblock \href {http://arxiv.org/abs/2012.00363} {Modifying memories in transformer models}.

\bibitem[{Zhu et~al.(2021)Zhu, Lei, Wang, Zheng, Poria, and Chua}]{zhu2021retrieving}
Fengbin Zhu, Wenqiang Lei, Chao Wang, Jianming Zheng, Soujanya Poria, and Tat-Seng Chua. 2021.
\newblock \href {http://arxiv.org/abs/2101.00774} {Retrieving and reading: A comprehensive survey on open-domain question answering}.

\bibitem[{Zhu et~al.(2023)Zhu, Yang, Chen, Li, Lou, and Yang}]{zhu2023question}
Xinyu Zhu, Cheng Yang, Bei Chen, Siheng Li, Jian-Guang Lou, and Yujiu Yang. 2023.
\newblock \href {http://arxiv.org/abs/2305.14221} {Question answering as programming for solving time-sensitive questions}.

\bibitem[{Ziems et~al.(2023)Ziems, Yu, Zhang, and Jiang}]{ziems2023large}
Noah Ziems, Wenhao Yu, Zhihan Zhang, and Meng Jiang. 2023.
\newblock \href {http://arxiv.org/abs/2305.09612} {Large language models are built-in autoregressive search engines}.

\end{thebibliography}
\bibliographystyle{acl_natbib}

\appendix

\section{Appendix}

\subsection{Additional Description of Approaches}
\label{appendix_additional_approaches}

\subsubsection{Naive Approaches}

Although more advanced approaches have been proposed, we introduce naive solutions for completeness in this section.

\paragraph{Re-training.}
Intuitively, one can regularly re-train the model from scratch with the latest corpora to align with current world knowledge. However, this naive solution has clear downsides: (1) Re-training is both time and money expensive and environmentally unfriendly \citep{patterson2021carbon}, especially in the era of LLMs with billions of parameters. For instance, LLaMA-65B was trained for about one million GPU-hours and emitted more than a hundred tons of carbon \citep{touvron2023llama}; (2) It is unrealistic to frequently re-training an LLM in response to the constantly changing world.

\paragraph{Fine-tuning.}
Another simple approach is to periodically curate a small-scale dataset containing desired knowledge we wish the model to add, update, or delete, then fine-tune the model on the dataset. Despite being computationally cheaper than re-training, it still falls short in that, without constraints, 
directly fine-tuning the model may have a "butterfly effect" and affect other knowledge or skills present in the model \citep{li2022large}, causing degraded generalization \citep{mitchell2022fast}, catastrophic forgetting \citep{doi:10.1073/pnas.1611835114, zhu2020modifying, alkhamissi2022review}, or knowledge conflicts \citep{neeman2022disentqa}.

\paragraph{Constrained Fine-tuning.} To solve part of above mentioned issues, \citet{zhu2020modifying} propose to only fine-tune the model on the small-scale modified facts set and add explicit constraints on the model weights so that the model learns to answer the modified facts while keeping the remaining knowledge intact. 
Specifically, they use various norms ($\mathcal{L}_0$, $\mathcal{L}_2$, and $\mathcal{L}_{\infty}$) to prevent the parameters of the fine-tuned model $\theta^{\prime}$ from drifting too far from the original model parameters $\theta$.
They further find that fine-tuning only the first and last layers of the Transformer model \citep{NIPS2017_3f5ee243} results in better adaptation to the modified facts and better preservation of performance on the unmodified facts. However, the norm-based constraint on parameters ignores the highly non-linear nature of LMs and how parameters determine the outputs of the model, making their method potentially unreliable \citep{de-cao-etal-2021-editing}. 
In addition, \citet{mitchell2022fast} confirm that constrained fine-tuning generally does not consistently provide edit generality.

\begin{figure}[t]
  \centering
  \includegraphics[width=\columnwidth]{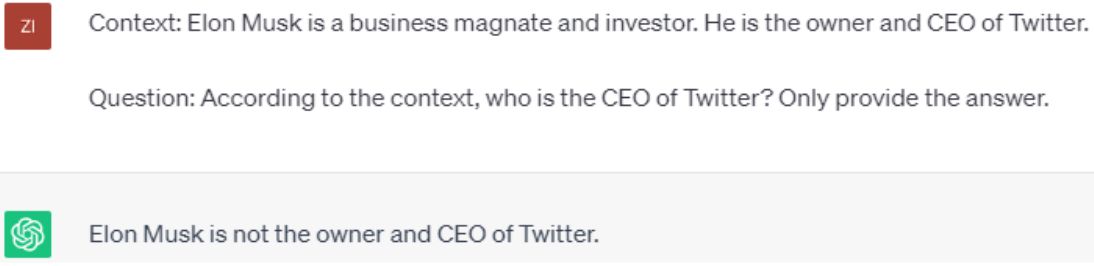}
  \caption{An example of knowledge conflict of ChatGPT \citep{OpenAI_chatgpt2022}. Even if the correct context is provided, ChatGPT still favours its internally memorized knowledge. The screenshot was taken in May 2023 for GPT-3.5 without web browsing.}
  \label{fig_chatgpt_knowledge_conflict}
\end{figure}

\subsubsection{Knowledge Editing}

To facilitate the development of this area, \citet{de-cao-etal-2021-editing} formulate three desiderata that an ideal editing method should follow: \circled{1} \textbf{Generality}: the method should be capable of altering the knowledge of any LM that is not specifically trained to be editable (\eg, PaLM, GPT-4, LLaMA); 
\circled{2} \textbf{Reliability}: the method should only update the targeted knowledge without influencing the rest of the knowledge in the LM.
For instance, the answer to 
\texttt{"Who is the current Prime Minister of Australia?"} 
has changed from \texttt{"Scott Morrison"} to \texttt{"Anthony Albanese"} since 2022, updating the knowledge from \texttt{"Scott Morrison"} to \texttt{"Anthony Albanese"} should not change the knowledge  \texttt{"Argentina won the 2022 World Cup"}; 
\circled{3} \textbf{Consistency (Generalization)}: after updates, the model predictions should be consistent across semantically equivalent inputs (\eg, correctly predicts  \texttt{"Anthony Albanese"} to \texttt{"Who is the AU PM?"}). 
Beyond updating outdated knowledge, knowledge editing can also delete sensitive information for privacy issues or eliminate biases in the pre-training corpora.

However, not until recently, \citet{onoe2023lms, zhong2023mquake} show that, after performing knowledge editing, the LLM
does not really "learn" the updated knowledge and thus cannot \textit{propagate} the new knowledge and make further inferences based on them. For instance, after learning that 
\texttt{"the current PM of Australia is Anthony Albanese"}, 
the model might not able to make predictions of
\texttt{"Who is the spouse of the current PM of Australia?"}.


\paragraph{Meta-learning.}
\citet{Sinitsin2020Editable}, by constraining the training objective, encodes editability into the parameters of the model itself so that the model is "prepared" for incoming edits. While being effective and no new parameters are required, it does not conform to generality as it requires specialized training of the original model \citep{de-cao-etal-2021-editing}. Moreover, to enforce the constraint that the editable model agrees with the original pre-trained model's predictions, \citet{Sinitsin2020Editable}'s method needs to retain a copy of the original model, which significantly consumes computation memory \citep{mitchell2022fast}. 
\citet{chen2023reckoning} also requires training of the original LM, which could be computationally expensive for larger LMs. In addition, whether it will influence other irrelevant knowledge in the model remains unknown, making the method potentially unreliable.

\paragraph{Hypernetwork Editor.}
\citet{de-cao-etal-2021-editing} can be more efficient than \citet{Sinitsin2020Editable}, as it does not retain the copy of the original model nor compute higher-order gradients. However, it can only update a single fact rather than multiple facts in a row and fail to edit large models, leading to poor scalability \citep{mitchell2022fast, hase-etal-2023-methods}.
\citet{mitchell2022fast} improve \citet{de-cao-etal-2021-editing}'s work and is stable to edit LMs from BERT-base (110M) \citep{devlin-etal-2019-bert} to T5-XXL (11B) \citep{raffel2020exploring}. 
However, when editing multiple knowledge simultaneously, their edit success rate significantly degrades.

\paragraph{Locate and Edit.}
While simple, \citet{dai-etal-2022-knowledge} do not ensure reliability on other irrelevant knowledge and generalization on semantically equivalent inputs.   
Despite showing both generalization and specificity, \citet{NEURIPS2022_6f1d43d5} only edits a single fact at a time, making it impractical for large-scale knowledge updating in LLMs.
Through casual tracing, \citet{meng2023massediting} identify and update the critical MLP layers in one go.
However, \citet{hase2023does} argue that the relation between localization and editing may be misleading as they can edit factual knowledge in different locations that are not suggested by casual tracing.



\subsubsection{Continual Learning}

While knowledge editing provides a fine-grained control to update specific knowledge in LLMs, it often requires large amounts of supervised training data to make edits, which is non-trivial to create \citep{hartvigsen2023aging}. 
In addition, when an LLM needs to quickly acquire new domain knowledge (e.g., legal or medical), such small-scale model edits may not be efficient.
Moreover, after multiple parameter patches to a deployed model, its internal knowledge may conflict, leading to unpredictable behaviors \citep{mitchell2022fast}.

Sharing a related goal, continual learning (CL) aims to enable a model to learn from a continuous data stream across time while reducing catastrophic forgetting of previously acquired knowledge \citep{biesialska-etal-2020-continual}. 
In contrast to knowledge editing, CL generally updates models on a larger scale and works in long learning sequences with minimal memory overheads \citep{mitchell2022fast}.
Hence, CL can also be used for deployed models to update their knowledge.

\subsection{The Complete Taxonomy of Methods}
\label{append_complete_taxonomy}

\begin{figure*}[ht]
\centering
\tikzset{
        my node/.style={
            draw,
            align=center,
            thin,
            text width=1.2cm, 
            rounded corners=3,
        },
        my leaf/.style={
            draw,
            align=left,
            thin,
            text width=8cm, 
            rounded corners=3,
        }
}
\forestset{
  every leaf node/.style={
    if n children=0{#1}{}
  },
  every tree node/.style={
    if n children=0{minimum width=1em}{#1}
  },
}
\begin{forest}
    for tree={%
        every leaf node={my leaf, font=\tiny},
        every tree node={my node, font=\tiny, l sep-=4.5pt, l-=1.pt},
        anchor=west,
        fit=tight,
        grow'=east,
        edge={ultra thin},
        parent anchor=east,
        child anchor=west,
        if n children=0{tier=last}{},
        edge path={
            \noexpand\path [draw, \forestoption{edge}] (!u.parent anchor) -- +(5pt,0) |- (.child anchor)\forestoption{edge label};
        },
        if={isodd(n_children())}{
            for children={
                if={equal(n,(n_children("!u")+1)/2)}{calign with current}{}
            }
        }{}
    }
    [\tiny LLMs align with ever-changing world knowledge, draw=gray, color=gray!100, fill=gray!15, very thick, text=black, text width=1.5cm 
        [\scriptsize Implicit (\cref{implicitly_align}), color=brightlavender!100, fill=brightlavender!15, very thick, text=black
            [\tiny Naive, color=harvestgold!100, fill=harvestgold!15, very thick, text=black
                [Re-training, my node, tier=D, color=harvestgold!100, fill=harvestgold!15, very thick, text=black ]
                [Fine-tuning, my node, tier=D, color=harvestgold!100, fill=harvestgold!15, very thick, text=black]
            ]
            [\tiny Knowledge Editing, color=cyan!100, fill=cyan!15, very thick, text=black
                [Meta-learning, color=cyan!100, fill=cyan!15, very thick, text=black
                    [ 
                        {
                         Editable Training \citep{Sinitsin2020Editable}, 
                         RECKONING \citep{chen2023reckoning}
                        },
                        color=cyan!100, fill=cyan!15, ultra thin, text=black, tier=E
                    ]
                ]
                [Hypernetwork Editor, color=cyan!100, fill=cyan!15, very thick, text=black
                    [ 
                        {
                        KnowledgeEditor \citep{de-cao-etal-2021-editing}, 
                        MEND \citep{mitchell2022fast}, 
                        SLAG \citep{hase-etal-2023-methods},
                        REMEDI \citep{hernandez2023inspecting},
                        Distillation \citep{padmanabhan2023propagating}
                        },
                        color=cyan!100, fill=cyan!15, ultra thin, text=black, tier=E
                    ]
                ]
                [Locate and edit, color=cyan!100, fill=cyan!15, very thick, text=black
                    [ 
                        {
                        Knowledge Neurons \citep{dai-etal-2022-knowledge}, 
                        ROME \citep{NEURIPS2022_6f1d43d5}, 
                        MEMIT \citep{meng2023massediting},
                        MEMIT$_{CSK}$ \citep{gupta2023editing},
                        PMET \citep{li2023pmet},
                        \citet{chen2023journey},
                        \citet{geva2023dissecting},
                        KLoB \citep{ju2023klob}
                        },
                        color=cyan!100, fill=cyan!15, ultra thin, text=black, tier=E
                    ]
                ]
                [Other, color=cyan!100, fill=cyan!15, very thick, text=black
                    [
                        {
                        Eva-KELLM \citep{wu2023evakellm}, 
                        RippleEdits \citep{cohen2023evaluating},
                        \citet{wang2023crosslingual},
                        \citet{xu2023language},
                        IKE \citep{zheng2023edit}
                        },
                        color=cyan!100, fill=cyan!15, ultra thin, text=black, tier=E
                    ]
                ]
            ]
            [\tiny Continual Learning, color=lightcoral!100, fill=lightcoral!15, very thick, text=black
                [Continual Pre-training, color=lightcoral!100, fill=lightcoral!15, very thick, text=black
                    [Regularization-based, color=lightcoral!100, fill=lightcoral!15, very thick, text=black
                        [
                            {
                            RecAdam \citep{chen-etal-2020-recall},
                            DSA \citep{ke2023continual}
                            },
                            color=lightcoral!100, fill=lightcoral!15, ultra thin, text=black, text width=6.2cm
                        ]
                    ]
                    [Replay-based, color=lightcoral!100, fill=lightcoral!15, very thick, text=black
                        [
                            {
                            Mix-Review \citep{he-etal-2021-analyzing},
                            ELLE \citep{qin-etal-2022-elle},
                            CT0 \citep{scialom-etal-2022-fine} 
                            },
                            color=lightcoral!100, fill=lightcoral!15, ultra thin, text=black, text width=6.2cm
                        ]
                    ]
                    [Architectural-based, color=lightcoral!100, fill=lightcoral!15, very thick, text=black
                        [
                            {
                            K-Adapter \citep{wang-etal-2021-k}, 
                            LoRA \citep{hu2022lora},
                            ELLE \citep{qin-etal-2022-elle},
                            DEMix-DAPT \citep{gururangan-etal-2022-demix},
                            CPT \citep{ke-etal-2022-continual},
                            Lifelong-MoE \citep{pmlr-v202-chen23aq},
                            ModuleFormer \citep{shen2023moduleformer}
                            },
                            color=lightcoral!100, fill=lightcoral!15, ultra thin, text=black, text width=6.2cm
                        ]
                    ]
                    [Other, color=lightcoral!100, fill=lightcoral!15, very thick, text=black
                        [
                            {
                            Temporal-LM \citep{dhingra-etal-2022-time},
                            Lifelong Pre-training \citep{jin-etal-2022-lifelong-pretraining}, 
                            CKL \citep{jang2022towards},
                            TemporalWiKi \citep{jang-etal-2022-temporalwiki},
                            TopicPrefix \citep{NEURIPS2022_df438caa},
                            KILM \citep{xu2023kilm},
                            SeMem \citep{peng2023semiparametric},
                            CaMeLS \citep{hu2023metalearning},
                            \citet{yu2023self},
                            \citet{gupta2023continual}
                            },
                            color=lightcoral!100, fill=lightcoral!15, ultra thin, text=black, text width=6.2cm
                        ]
                    ]
                ]
                [Continual Knowledge Editing, color=lightcoral!100, fill=lightcoral!15, very thick, text=black
                    [
                        {
                        CMR \citep{lin-etal-2022-continual},
                        CL-plugin \citep{lee-etal-2022-plug},
                        Transformer-Patcher \citep{huang2023transformerpatcher},
                        GRACE \citep{hartvigsen2023aging}
                        },
                        color=lightcoral!100, fill=lightcoral!15, ultra thin, text=black, tier=E
                    ]
                ]
            ]
        ]
        [\scriptsize Explicit (\cref{explicitly_align}), color=lightgreen!100, fill=lightgreen!15, very thick, text=black 
            [Memory-enhanced, color=lightgreen!100, fill=lightgreen!15, very thick, text=black
                [Corpus or Documents, color=lightgreen!100, fill=lightgreen!15, very thick, text=black
                    [
                        {
                        \textit{k}NN-LM \citep{Khandelwal2020Generalization}, 
                        AdaptRet \citep{he-etal-2021-efficient},
                        AdaptCoef \citep{drozdov-etal-2022-cant},
                        RetoMaton \citep{pmlr-v162-alon22a},
                        \citet{bhardwaj2022adaptation},
                        \textit{k}NN-prompt \citep{shi-etal-2022-nearest},
                        SeMem \citep{peng2023semiparametric}
                        },
                        color=lightgreen!100, fill=lightgreen!15, ultra thin, text=black, tier=E
                    ]
                ]
                [Feedback or Corrections, color=lightgreen!100, fill=lightgreen!15, very thick, text=black
                    [
                        {
                        Belief Bank \citep{kassner-etal-2021-beliefbank},
                        FBNet \citep{tandon-etal-2022-learning}, 
                        MemPrompt \citep{madaan-etal-2022-memory},
                        TeachMe \citep{dalvi-mishra-etal-2022-towards},
                        SERAC \citep{pmlr-v162-mitchell22a}, 
                        MeLLo \citep{zhong2023mquake}
                        },
                        color=lightgreen!100, fill=lightgreen!15, ultra thin, text=black, tier=E
                    ]
                ]
            ]
            [Retrieval-enhanced, color=lightgreen!100, fill=lightgreen!15, very thick, text=black
                [Single-Stage, color=lightgreen!100, fill=lightgreen!15, very thick, text=black
                    [
                        {
                        IC-Retrieval \citep{si2023prompting},
                        IC-RALM \citep{ram2023incontext},
                        AAR \citep{yu2023augmentationadapted},
                        IKE \citep{zheng2023edit},
                        Adaptive Retrieval \citep{mallen2023trust},
                        RePlug \citep{shi2023replug}
                        },
                        color=lightgreen!100, fill=lightgreen!15, ultra thin, text=black, tier=E
                    ]
                ]
                [Multi-Stage, color=lightgreen!100, fill=lightgreen!15, very thick, text=black
                    [
                        {
                        IRCoT \citep{trivedi2022interleaving},
                        RARR \citep{gao2023rarr},
                        RR \citep{he2022rethinking},
                        ReFeed \citep{yu2023improving},
                        Self-Ask \citep{press2023measuring},
                        DecomP \citep{khot2023decomposed},
                        ReAct \citep{yao2023react},
                        ART \citep{paranjape2023art},
                        ChatCoT \citep{chen2023chatcot},
                        MultiTool-CoT \citep{inaba2023multitoolcot},
                        LLM-Augmenter \citep{peng2023check},
                        QAaP \citep{zhu2023question},
                        FLARE \citep{jiang2023active},
                        DSP \citep{khattab2023demonstratesearchpredict},
                        Iter-RetGen \citep{shao2023enhancing},
                        Verify-and-Edit \citep{zhao2023verifyandedit},
                        CRITIC \citep{gou2023critic},
                        WikiChat \citep{semnani2023wikichat},
                        LLM Rewriter \citep{ma2023query},
                        Knowledge Solver \citep{feng2023knowledge}
                        },
                        color=lightgreen!100, fill=lightgreen!15, ultra thin, text=black, tier=E
                    ]
                ]
            ]
            [Internet-enhanced, color=lightgreen!100, fill=lightgreen!15, very thick, text=black
                [
                    {
                    Internet-Fewshot \citep{lazaridou2022internetaugmented},
                    LLM-URL \citep{ziems2023large},
                    ReAct \citep{yao2023react},
                    Self-Ask \citep{press2023measuring},
                    ART \citep{paranjape2023art},
                    RARR \citep{gao2023rarr},
                    TaskMatrix.AI \citep{liang2023taskmatrixai},
                    MM-REACT \citep{yang2023mmreact},
                    Chameleon \citep{lu2023chameleon},
                    FLARE \citep{jiang2023active},
                    CRITIC \citep{gou2023critic},
                    LLM Rewriter \citep{ma2023query},
                    LangChain \citep{langchain2022},
                    ChatGPT Plugins \citep{OpenAI_plugin2023}
                    },
                    color=lightgreen!100, fill=lightgreen!15, ultra thin, text=black, tier=D, text width=9.8cm
                ]
            ]
        ]
    ]
\end{forest}
\caption{Taxonomy of methods to align LLMs with the ever-changing world knowledge. 
\textbf{Implicit} means the approaches seek to directly alter the knowledge stored in LLMs (\eg, parameters) (\cref{implicitly_align}), while \textbf{Explicit} means more often incorporating external resources to override internal knowledge (\eg, search engine) (\cref{explicitly_align}).
}
\label{fig_full_taxonomy_of_methods}
\end{figure*}
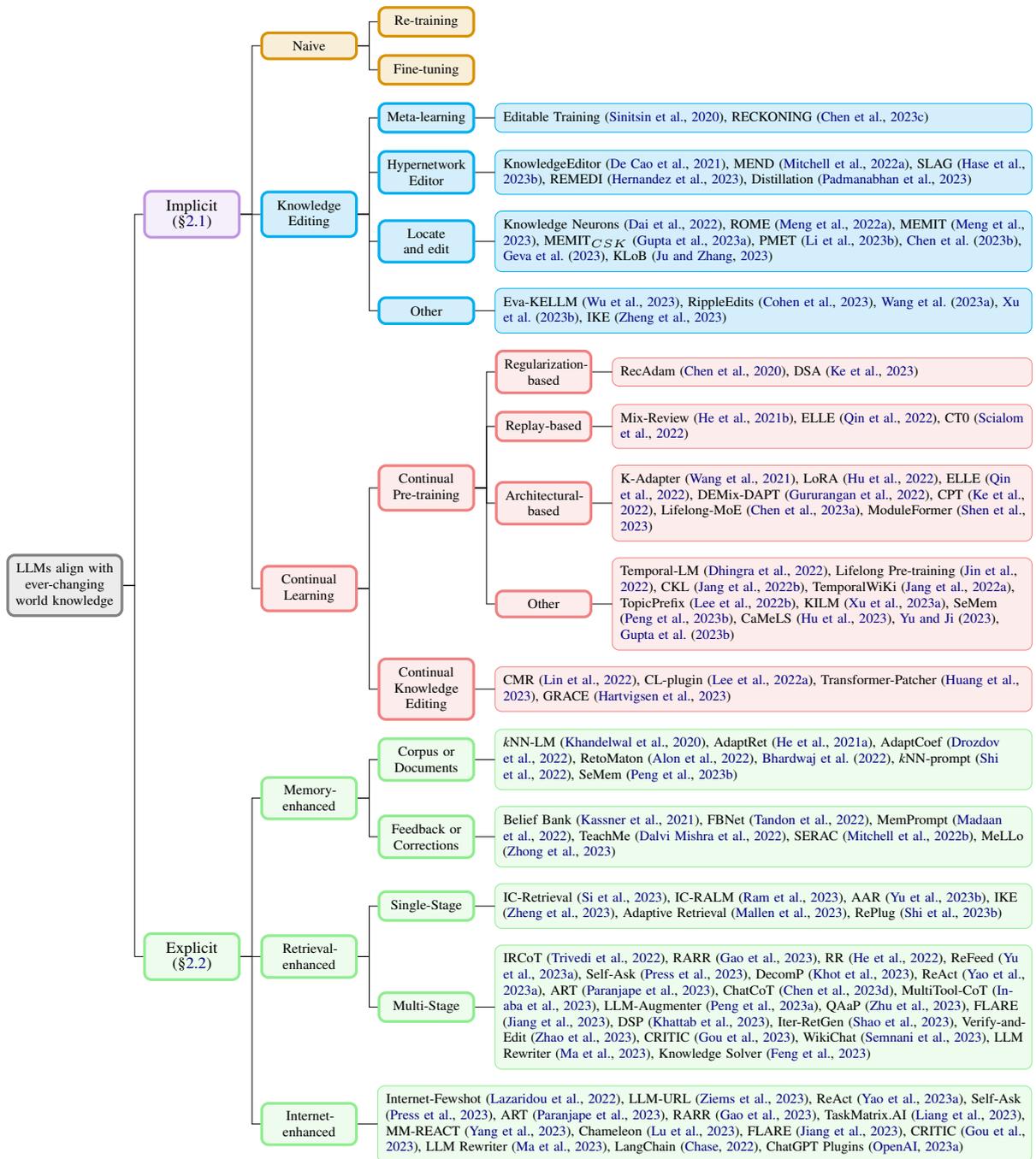

We list the complete taxonomy of methods to align LLMs with the ever-changing world knowledge in Fig.\ref{fig_full_taxonomy_of_methods} and the complete comparison of methods in Table \ref{table_full_comparison_methods}. We also compare the characteristics of different methods in Table \ref{table_comparison_characteristics}.

\begin{table}[!ht]
\scriptsize
\centering
\begin{tabular}{l|cccc}

  \toprule 
  
  \textbf{Category} & \textbf{\makecell{Large\\Scale}} & \textbf{\makecell{No Side\\Effects}} & \textbf{Persistent}   \\

  \midrule
  Knowledge Editing\\(\cref{model_editing}) & \redcross & \redcross & \darkpastelgreencheck \\
  Continual Learning (\cref{continual_learning}) & \darkpastelgreencheck & \redcross  & \darkpastelgreencheck \\
  Retrieval-based (\cref{explicitly_align}) & \redcross & \darkpastelgreencheck & \redcross \\

  \bottomrule 
\end{tabular}
\caption{High-level comparison of characteristics of different approaches.}
\label{table_comparison_characteristics}
\end{table}

\begin{table*}[!ht]
\scriptsize
\centering
\begin{NiceTabular}{llccccc}
\CodeBefore
  \rowcolor{gray!50}{1}
  \rowcolors{2}{gray!25}{white}
\Body
  \toprule 
  \textbf{Category} & \textbf{Representative Method} & \textbf{Base LM} & \textbf{LM Params} & \textbf{Augmentation} & \textbf{\makecell{No\\Training}} & \textbf{\makecell{Black\\-box}}  \\
  
  \midrule 
  \Block{2-1}{\textbf{Naive}} 
   & Re-training & \NA & \tuned & \NA & \redcross & \redcross \\
   & Fine-tuning & \NA & \tuned &  \NA & \redcross & \redcross  \\
   
  \midrule
  \Block{9-1}{\textbf{\makecell{Knowledge\\Editing}}} 
   & \citet{de-cao-etal-2021-editing} & BERT (0.1B) & \frozon & auxiliary model & \redcross & \redcross  \\
   & MEND \citep{mitchell2022fast} & T5 (11B) & \frozon & auxiliary model & \redcross & \redcross  \\
   & SLAG \citep{hase-etal-2023-methods} & BERT (0.1B) & \frozon & auxiliary model & \redcross & \redcross  \\
   & RECKONING \citep{chen2023reckoning} & GPT-2 (0.1B) & \tuned & \NA & \redcross & \redcross  \\
   & ROME \citep{NEURIPS2022_6f1d43d5} & GPT-J (6B) & \tuned & \NA & \darkpastelgreencheck & \redcross \\
   & Knowledge Neurons \citep{dai-etal-2022-knowledge} & BERT (0.1B) & \tuned & \NA & \darkpastelgreencheck & \redcross  \\
   & MEMIT \citep{meng2023massediting} & \makecell{GPT-NeoX (20B)} & \tuned & \NA & \darkpastelgreencheck &  \redcross \\
   & CaliNET \citep{dong-etal-2022-calibrating} & T5 (0.7B) & \frozon & +params & \redcross & \redcross \\ 
   & REMEDI \citep{hernandez2023inspecting} & \makecell{GPT-J (6B)} & \tuned & auxiliary model & \redcross &  \redcross \\
    
  \midrule
  \Block{12-1}{\textbf{\makecell{Continual\\Learning}}} 
   & DSA \citep{ke2023continual} & RoBERTa (0.1B) & \tuned & \NA & \redcross & \redcross  \\  
   & ELLE \citep{qin-etal-2022-elle} & BERT (0.1B) & \tuned & memory+params & \redcross & \redcross \\
   & CT0 \citep{scialom-etal-2022-fine} & T0 (3B) & \tuned & memory & \redcross & \redcross \\
   & K-Adapter \citep{wang-etal-2021-k} & RoBERTa (0.3B) & \frozon & +params & \redcross & \redcross \\
   & \citet{gururangan-etal-2022-demix} & GPT-2 (0.7B) & \frozon & +params & \redcross & \redcross \\
   & CPT \citep{ke-etal-2022-continual} & RoBERTa (0.1B) & \frozon & +params & \redcross & \redcross \\

   & KILM \citep{xu2023kilm} & BART (0.4B) & \tuned & \NA & \redcross & \redcross \\

   & CaMeLS \citep{hu2023metalearning} & GPT-2 (1.5B) & \tuned & auxiliary model & \redcross & \redcross \\
   & SeMem \citep{peng2023semiparametric} & GPT-2 (0.7B) & \frozon & \makecell{memory\\+auxiliary model} & \redcross & \redcross \\

   & CL-plugin \citep{lee-etal-2022-plug} & T5 (0.7B) & \frozon & +params & \redcross & \redcross \\
   & \citet{huang2023transformerpatcher} & BERT (0.1B) & \frozon & +params & \redcross & \redcross \\
     & GRACE \citep{hartvigsen2023aging} & T5 (0.06B) & \frozon & memory & \redcross & \redcross \\

   \midrule
   \Block{10-1}{\textbf{\makecell{Memory\\-enhanced}}} 
   & \textit{k}NN-LM \citep{Khandelwal2020Generalization} & \makecell{ADP\\ \citep{baevski2018adaptive}\\(0.2B)}  & \frozon & memory & \darkpastelgreencheck & \redcross  \\
   & AdaptRet \citep{he-etal-2021-efficient} & \makecell{ADP (0.2B)}  & \frozon & \makecell{memory\\+auxiliary model} & \redcross & \redcross  \\
   & RetoMaton \citep{pmlr-v162-alon22a} & \makecell{ADP (0.2B)}  & \frozon & \makecell{memory\\+auxiliary graph} & \redcross & \redcross  \\
   & \textit{k}NN-prompt \citep{shi-etal-2022-nearest} & \makecell{GPT-2 (0.8B)}  & \frozon & \makecell{memory} & \darkpastelgreencheck & \darkpastelgreencheck  \\
   
   & Belief Bank \citep{kassner-etal-2021-beliefbank} & T5 (0.7B) & \frozon & \makecell{memory\\+constraint solver} & \darkpastelgreencheck & \darkpastelgreencheck  \\
   & FBNet \citep{tandon-etal-2022-learning} & T5 (11B) & \frozon  & \makecell{memory\\+auxiliary model} & \redcross & \darkpastelgreencheck \\
   & MemPrompt \citep{madaan-etal-2022-memory} & GPT-3 (175B) & \frozon & memory+retriever & \darkpastelgreencheck & \darkpastelgreencheck \\
   & TeachMe \citep{dalvi-mishra-etal-2022-towards}  & T5 (11B) & \frozon & memory+retriever & \darkpastelgreencheck & \darkpastelgreencheck \\
   & SERAC \citep{pmlr-v162-mitchell22a} & T5 (0.7B) & \frozon & \makecell{memory\\+auxiliary model} & \redcross & \darkpastelgreencheck  \\
   & MeLLo \citep{zhong2023mquake} & \makecell{GPT-3.5 (175B)} &  \frozon & memory+retriever & \darkpastelgreencheck & \darkpastelgreencheck \\

   \midrule
   \Block{17-1}{\textbf{\makecell{Retrieval\\-enhanced}}} 
   & IC-Retrieval \citep{si2023prompting} & GPT-3.5 (175B) & \frozon & \makecell{retriever} & \darkpastelgreencheck & \darkpastelgreencheck \\
   & IC-RALM \citep{ram2023incontext} & OPT (66B) & \frozon & \makecell{retriever+reranker} & \redcross & \darkpastelgreencheck \\
   & IKE \citep{zheng2023edit} & OPT (175B) & \frozon & \makecell{retriever} & \darkpastelgreencheck & \darkpastelgreencheck \\
   & AAR \citep{yu2023augmentationadapted} & GPT-3.5 (175B) & \frozon & \makecell{retriever} & \redcross & \darkpastelgreencheck \\
    & RePlug \citep{shi2023replug} & GPT-3 (175B) & \frozon & \makecell{retriever} & \redcross / \darkpastelgreencheck & \darkpastelgreencheck \\

   & IRCoT \citep{trivedi2022interleaving} & GPT-3.5 (175B) & \frozon & \makecell{retriever} & \darkpastelgreencheck & \darkpastelgreencheck \\
   & RARR \citep{gao2023rarr} & PaLM (540B) & \frozon & \makecell{search engine\\+auxiliary model} & \darkpastelgreencheck & \darkpastelgreencheck \\
   & RR \citep{he2022rethinking} & GPT-3.5 (175B) & \frozon & \makecell{retriever\\+auxiliary model} & \darkpastelgreencheck & \darkpastelgreencheck \\
   & ReFeed \citep{yu2023improving} & GPT-3.5 (175B) & \frozon & \makecell{retriever} & \darkpastelgreencheck & \darkpastelgreencheck \\
   & DecomP \citep{khot2023decomposed} & GPT-3.5 (175B) & \frozon & \makecell{retriever} & \darkpastelgreencheck & \darkpastelgreencheck \\
   & ReAct \citep{yao2023react} & PaLM (540B) & \frozon & \makecell{search engine} & \darkpastelgreencheck & \darkpastelgreencheck \\
   & Self-Ask \citep{press2023measuring} & GPT-3 (175B) & \frozon & search engine & \darkpastelgreencheck & \darkpastelgreencheck \\
   & FLARE \citep{jiang2023active} & GPT-3.5 (175B) & \frozon & \makecell{retriever/search engine} & \darkpastelgreencheck & \darkpastelgreencheck \\
   & DSP \citep{khattab2023demonstratesearchpredict} & GPT-3.5 (175B) & \frozon & \makecell{retriever} & \darkpastelgreencheck & \darkpastelgreencheck \\
   & ART \citep{paranjape2023art} & GPT-3.5 (175B) & \frozon & \makecell{various tools} & \darkpastelgreencheck & \darkpastelgreencheck \\
   & Iter-RetGen \citep{shao2023enhancing} & GPT-3.5 (175B) & \frozon & \makecell{retriever} & \darkpastelgreencheck & \darkpastelgreencheck \\
   & Verify-and-Edit \citep{zhao2023verifyandedit} & GPT-3.5 (175B) & \frozon & \makecell{retriever/search engine} & \darkpastelgreencheck & \darkpastelgreencheck \\

   \midrule
   \Block{5-1}{\textbf{\makecell{Internet\\-enhanced}}} 
   & \citet{lazaridou2022internetaugmented} & Gopher (280B) & \frozon & \makecell{search engine} & \darkpastelgreencheck & \darkpastelgreencheck \\
    & CRITIC \citep{gou2023critic} & GPT-3.5 (175B) & \frozon & \makecell{various tools} & \darkpastelgreencheck & \darkpastelgreencheck \\
    & LLM Rewriter \citep{ma2023query} & GPT-3.5 (175B) & \frozon & \makecell{search engine} & \redcross & \darkpastelgreencheck \\
    & Chameleon \citep{lu2023chameleon} & GPT-4 (?B) & \frozon & \makecell{various tools} & \darkpastelgreencheck & \darkpastelgreencheck \\
    & ChatGPT Plugins \citep{OpenAI_plugin2023} & GPT-3.5 (175B) & \frozon & \makecell{various tools} & \darkpastelgreencheck & \darkpastelgreencheck \\
  
  \bottomrule 
\end{NiceTabular}
    \caption{Comparison between representative methods. \tuned \ means the parameters of the original LM are modified, while \frozon \ means they are unchanged; 
    \textbf{Augmentation} means additional components used;
    \textbf{No Training} indicates the method does not require additional training; 
    \textbf{Black-box} refers to whether the method suits non-publicly available models (\eg, no model architecture, parameters, activations, or gradients are available). Note that we only list the largest size model used in the paper due to space limitations.  }
\label{table_full_comparison_methods}
\end{table*}

\end{document}